\documentclass[final,5p,times,twocolumn,number]{elsarticle}
 
\usepackage{xcolor} 
\usepackage{soul}   

\usepackage{tabularx} 
\usepackage{graphicx} 
\usepackage{amssymb} 
\usepackage{booktabs}   
\usepackage{amsmath} 
\usepackage{hyperref}  
\usepackage{cleveref}
\usepackage{cite}
\usepackage{amsfonts}
\usepackage{algorithmic}
\usepackage{textcomp}
\usepackage[table]{xcolor}
\usepackage{color,colortbl}
\usepackage{wrapfig}
\usepackage{array}
\usepackage{bbding}
\usepackage{arydshln}
\usepackage{CJKutf8}
\usepackage{multirow}
\usepackage{enumitem} 
\usepackage{makecell}
\usepackage{bbm}

\newcommand{\s}[1]{}
\newcommand\greenp[1]{\textcolor{darkergreen}{}}
\usepackage[linesnumbered,ruled,vlined]{algorithm2e}

\newcolumntype{I}{!{\vrule width 1.2pt}}
\newlength\savedwidth
\newcommand\whline{\noalign{\global\savedwidth\arrayrulewidth
                            \global\arrayrulewidth 1.5pt}
                   \hline
                   \noalign{\global\arrayrulewidth\savedwidth}}
\newlength\savewidth

\newcommand{\tocite}[1]{{\textcolor{red}{TOCITE}}}

\newcommand{\ud}[1]{\textcolor{mypink}{#1}}

\definecolor{mypink}{rgb}{.99,.91,.95}
\definecolor{mygreen}{RGB}{107,147,147}
\definecolor{lgreen}{HTML}{00b8a9}
\definecolor{mygray}{HTML}{eeeeee}
\definecolor{myred}{HTML}{fae4df}
\definecolor{myyellow}{HTML}{FCF0C7}
\definecolor{darkergreen}{RGB}{21, 152, 56}
\definecolor{newcolor}{rgb}{.8,.349,.1}
\definecolor{mypink}{rgb}{0,0,0}

\journal{Knowledge-Based Systems} 

\begin{document}

\begin{frontmatter}

\title{DBHiCo: Dual-Branch Hierarchical Contrastive Learning for MRI Image Classification}







\author[inst1,inst2]{Chengsheng Zhang}
\ead{zhangcs66@mail.ustc.edu.cn}

\author[inst1,inst2]{Wei Li}
\ead{lwzkd@mail.ustc.edu.cn}

\author[inst1,inst2]{Chenghao Sun}
\ead{chsun@mail.ustc.edu.cn}

\author[inst3]{Jian Gong}
\ead{gongjian88@vip.163.com}
  
\author[inst1,inst2]{Xinmei Tian\corref{cor1}}
\ead{xinmei@ustc.edu.cn}
 
\cortext[cor1]{Corresponding author.}

\affiliation[inst1]{
            organization={University of Science and Technology of China},
            city={Hefei},
            postcode={230026}, 
            state={Anhui},
            country={China}}

\affiliation[inst2]{organization={Institute of Artificial Intelligence},
            addressline={Hefei Comprehensive National Science Center}, 
            city={Hefei},
            postcode={230000}, 
            state={Anhui},
            country={China}}

\affiliation[inst3]{organization={Beijing Tiantan Hospital, Capital Medical University},
            addressline={Beijing Neurosurgical Institute}, 
            postcode={100071}, 
            state={Beijing},
            country={China}}

\begin{abstract}
Automated Magnetic Resonance Imaging (MRI) image classification provides a crucial non-invasive approach for neurological diagnosis, yet it is fundamentally challenged by inherent acquisition noise and high inter-subject structural similarity, which obscure subtle pathological variations, thereby limiting the performance of deep neural networks.
While contrastive learning (CL) offers a promising mitigation strategy, existing paradigms rely on rigid binary partitions that ignore negative sample heterogeneity, and they fail to capture the cross-view structural consistency vital for fine-grained anatomical alignment. 
To address these barriers, we propose Dual-Branch Hierarchical Contrastive Learning (DBHiCo), a novel framework explicitly designed to collaboratively enforce structural consistency and semantic discrimination. 
Specifically, DBHiCo leverages a natural anatomical multi-view pair—original and skull-stripped MRI scans—processing them through a dual-branch architecture to seamlessly integrate holistic structural robustness with localized intracranial details. 
Crucially, we introduce a hierarchical contrastive loss that discards conventional binary constraints in favor of a margin-ranked similarity hierarchy. This mechanism explicitly orders positive and negative pairs based on their structural and semantic difficulty, thereby amplifying the fine-grained semantic discriminability of the learned representations. 
Extensive experiments across three diverse neurological datasets demonstrate that DBHiCo achieves state-of-the-art performance, consistently outperforming existing MRI multi-view representation learning and advanced contrastive fine-tuning methods.
\end{abstract}


\end{frontmatter}


\vspace{-7pt}
\section{Introduction}\label{sec:intro}

Magnetic Resonance Imaging (MRI) serves as a cornerstone non-invasive tool for capturing internal structural and functional information, making it indispensable for identifying neurological abnormalities~\citep{katti2011magnetic,mousavian2021depression,hojjati2018predicting}. 
While computer-aided diagnosis (CAD) systems have profoundly alleviated the manual interpretation burden of MRI places on radiologists~\citep{bouhafra2025deep,alexander2022mandating,kwee2021workload}, empowering the efficient diagnosis of neurological disorders such as pediatric posterior fossa tumors~\citep{sun2024expert,reddy2024brain,singh2024evaluating}, Alzheimer’s disease~\citep{marcus2007open,kunanbayev2024training,liu2025simcmc,cui2025novel}, and Attention-Deficit Hyperactivity Disorder~\citep{adhd2012adhd,kuang2014classification,qureshi2016classification,shi2025transformer}, the core deep neural networks (DNNs)~\citep{dosovitskiy2020image, he2016deep,liu2021swin,touvron2021training} driving these systems face unique challenges: 
Unlike natural images, MRI acquisitions inevitably introduce system noise~\citep{liu2025simcmc} that causes random intensity fluctuations—which DNNs may misinterpret as false diagnostic features—and exhibit high inter-subject similarity~\citep{sebenius2025structural} that obscures subtle pathological variations. 
Consequently, direct transfer of large-scale pre-trained knowledge through standard supervised learning frequently induces overfitting. 
The network tends to memorize random intensity fluctuations rather than extracting robust, disease-relevant anatomical patterns~\citep{khan2023transfer,srinivas2022deep,sharma2023brain,rajput2024transfer,mehmood2021transfer}.



To mitigate this bottleneck, recent approaches have increasingly adopted multi-view representation learning paradigms~\citep{valanarasu2023disruptive,wu2024voco,zhuang2025mim,rui2025multi}. 
By applying contrastive learning (CL) across multiple distinct views derived from a single MRI structure (\textit{e.g.}, skull-stripped volume)—such as through spatial masking or multi-level cropping~\citep{chen2020simple,taleb2017self}—these methods effectively compel the network to extract robust features via pre-training. 
Despite their effectiveness, these methods overlook multi-view construction from an anatomical structural perspective. 
In standard neuroimaging pipelines, an exceptionally natural structural multi-view pair is inherently facilitated by the skull-stripping pre-processing step~\citep{ridha2024improves,acosta2008impact,lakshmi2014noise}. 
Notably, original MRI scans preserve holistic anatomical boundaries and soft-tissue continuity, whereas skull-stripped scans isolate intracranial parenchymal details~\citep{kalavathi2016methods,digregorio2021intracranial,isensee2019automated}. 
Therefore, these two views provide highly complementary information, \textit{i.e.}, the anatomical structural integrity and the salience of tissue-level details, which are critical for accurate diagnosis.

\begin{figure}[!ht]
    \begin{center}
    \vspace{-5pt}
	\includegraphics[width=1\linewidth]{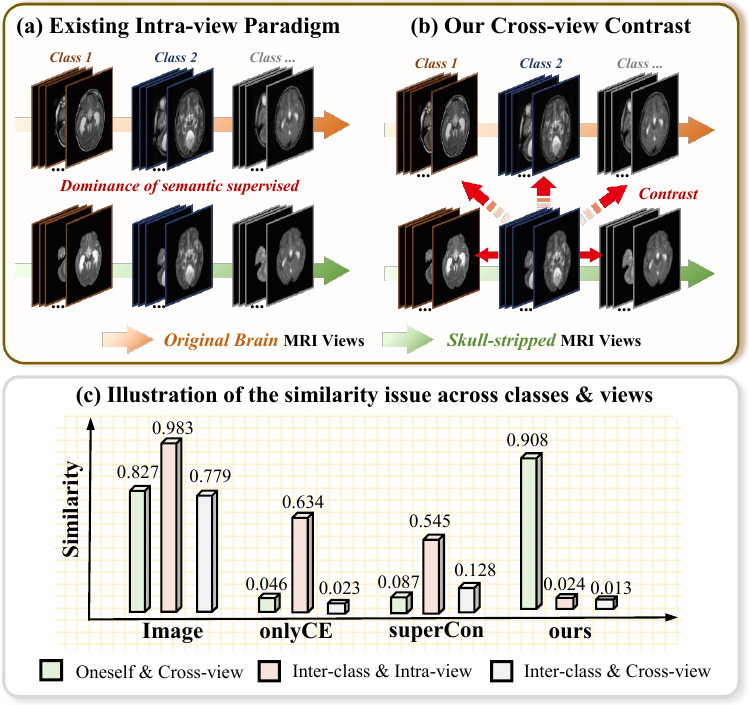}
    \end{center}
    \vspace{-5pt}
    \caption{
        Motivation of DBHiCo. 
        (a)-(b) Comparison between the semantic supervised paradigms and ours. Intra-views supervision provides insufficient learning of structural relationships, whereas our method explicitly enforces cross-view and semantic contrastive learning.
        (c) Similarity analysis across views and classes. 
        MRI images exhibit high similarity issue within the same view, while previous method (e.g., SuperCon ~\citep{khosla2020supervised}) disproportionately emphasize semantic discrimination but ignoring structural consistency learning.
    }
    \label{fig:intro}
    \vspace{-10pt}
\end{figure}


However, existing paradigms fail to fully unlock the potential of such complementary cross-view data for downstream clinical tasks. 
Specifically, current multi-view pre-training methods heavily rely on the curation of massive MRI cohorts, and their self-supervised proxy objectives are often disconnected from specific diagnostic goals, leading to sub-optimal transferability. 
To overcome this task disconnect, an intuitive alternative is to directly optimize representations on the downstream task via standard semantic supervision~\citep{chen2020simple,zhu2022balanced,du2024probabilistic}. 
Nevertheless, these approaches focus predominantly on global semantic discrimination (Figure~\hyperref[fig:intro]{\ref*{fig:intro}-a}), thereby neglecting the cross-view \textbf{\textit{structural consistency}} required to align fine-grained anatomical features. 
Furthermore, they typically employ a rigid binary partition—pulling same-class views together while pushing all other instances apart uniformly~\citep{khosla2020supervised}. 
This simplistic strategy ignores the inherent heterogeneity among negative samples (\textit{e.g.}, varying degrees of intra-class vs. inter-class structural variations), treating them equally and thereby yielding a latent space that lacks the granularity necessary for precise \textbf{\textit{semantic discrimination}} in complex brain structures (Figure~\hyperref[fig:intro]{\ref*{fig:intro}-c}).


To address these barriers, we propose the Dual-Branch Hierarchical Contrastive Learning (DBHiCo) method. 
Specifically, DBHiCo is explicitly designed to collaboratively enforce structural consistency and semantic discrimination during training (Figure~\hyperref[fig:intro]{\ref*{fig:intro}-b}). 
First, we introduce a dual-branch architecture that extracts and fuses features from paired original and skull-stripped MRI scans in parallel, integrating holistic structural robustness with localized semantic richness. 
Second, to rectify the limitations of binary contrastive targets, we propose a novel hierarchical contrastive loss ($\mathcal{L}_{\text{HiCo}}$). 
In contrast to a uniform penalty, $\mathcal{L}_{\text{HiCo}}$ establishes a strict similarity ranking constraint across positive and varying-difficulty negative pairs. By explicitly modeling these hierarchical disparities via margin parameters driven by both structural (cross-view) and semantic (cross-class) conditions, the network is forced to capture subtle anatomical nuances. 
Finally, by jointly optimizing this hierarchical contrastive objective with the downstream classification task, DBHiCo directly tailors the learned representations to clinical requirements.

We experimentally validate DBHiCo on three representative neurological disorders benchmark: our collected Tiantan dataset, public OASIS-1~\citep{marcus2007open} dataset, and ADHD-200~\citep{adhd2012adhd} competition dataset.
DBHiCo achieves superior performance on various baseline model architecture, outperforming existing cross-view representation learning and supervised contrastive learning studies. 
 
Our principal contributions can be summarized as follows.
  
\begin{itemize}
    \item We present DBHiCo, a novel multi-view contrastive learning framework tailored for brain MRI classification. By coupling dual-branch feature fusion with cross-view similarity constraints, DBHiCo effectively overcomes the inter-subject similarity bottlenecks inherent in neuroimaging analysis.
    
    \item We propose a hierarchical contrastive loss function ($\mathcal{L}_{\text{HiCo}}$) that discards the prior binary contrast in favor of a margin-ranked similarity hierarchy. This explicitly orders positive and negative pairs based on structural and semantic difficulty, amplifying fine-grained semantic discrimination.
    
    \item Extensive experiments demonstrate that DBHiCo achieves state-of-the-art performance, consistently outperforming existing specialized MRI multi-view representation learning and advanced contrastive fine-tuning methods.
\end{itemize}

\section{Related works}\label{eq:relatedworks}

Automated MRI image classification provides medical professionals with a non-invasive tool for diagnosing neurological disorders. Recently, CL has driven significant advancements in this domain through diverse representation optimization objectives. 
In this section, we first introduce existing multi-view representation learning methods tailored for MRI. The limitations inherent in these self-supervised pre-training paradigms motivate our exploration into applying CL directly to downstream clinical tasks. Furthermore, recognizing that conventional CL approaches often struggle to capture the intricate anatomical structures of the brain, we subsequently discuss the literature that inspires our proposed hierarchical contrastive design.

\subsection{MRI Multi-View Representation Learning.} 

Numerous studies are built upon the insight that integrating multi-view MRI data enables a more comprehensive characterization of brain structures and pathological features~\citep{talai2021utility}. 
Broadly, existing multi-view strategies can be categorized into two main paradigms. 
The first paradigm derives multiple views from distinct spatial regions of a single MRI volume. These methods typically employ techniques such as 3D volume masking (\textit{e.g.}, general~\citep{chen2023masked}, localized~\citep{valanarasu2023disruptive}, and global-local~\citep{zhuang2025advancing}) or spatial cropping (\textit{e.g.}, random~\citep{wu2024voco} and multi-level~\citep{zhuang2025mim}) to formulate self-supervised pretext tasks. 
The second paradigm leverages different MRI modalities as distinct views, driving model pre-training through cross-modal reconstruction~\citep{wang2024spatial,zhang2025deep,zeng2024missing}, multi-modal contrastive learning~\citep{liu2022joint,li2024multimodality,rui2025multi}, or puzzle-solving objectives~\citep{taleb2017self}. 

Despite their empirical success, these multi-view pre-training paradigms suffer from notable limitations: they heavily rely on the curation of exceptionally large-scale MRI pre-training cohorts, and their self-supervised proxy objectives are often disconnected from the specific goals of downstream clinical tasks, leading to sub-optimal transferability. 
In contrast, our DBHiCo effectively mitigates these bottlenecks. By introducing a novel cross-view strategy (Original vs. Skull-stripped) tailored to the intrinsic anatomical characteristics of MRI data, DBHiCo directly optimizes representations for downstream tasks. As validated in our experiments (see Section~\ref{sec:ablation}), this targeted approach achieves superior performance in capturing both structural and semantic nuances without the burden of massive pre-training.

\subsection{Contrastive Learning for Classification Tasks}

CL aims to construct a feature space that encodes semantic similarity by pulling positive sample pairs closer while pushing negative pairs apart. 
As a representative self-supervised CL framework, SimCLR~\citep{chen2020simple} leverages extensive data augmentations to introduce sufficient variations, enabling the network to learn robust and invariant representations. 
Building upon this foundation, SuperCon~\citep{khosla2020supervised} explicitly incorporates supervisory signals into the contrastive objective, yielding highly discriminative features when class labels are available. This paradigm shift toward supervised CL has significantly enhanced representation learning on labeled data, empowering researchers to tailor custom loss designs for specific data characteristics, particularly in data-scarce and class-imbalanced scenarios~\citep{zhu2022balanced,du2024probabilistic,zhang2023ecl,cai2024bpaco}. 

However, the predominant focus of these standard methods on global semantic discrimination limits their efficacy when applied to structure-rich MRI scans. Medical imaging requires a nuanced understanding of fine-grained anatomical details alongside global category semantics. To bridge this gap, our proposed DBHiCo introduces a novel hierarchical CL framework designed to achieve the integrated learning of structural consistency and semantic discrimination.

\section{Methodology}\label{sec:method}
This section provides an overview of our proposed DBHiCo method. Firstly, the overall framework of the DBHiCo method is introduced in Section~\ref{sec:framework}. Then, the process of constructing positive and negative pairs is presented in Section~\ref{sec:construct_pairs}. Finally, the hierarchical contrastive loss of the DBHiCo method is introduced in Section~\ref{sec:hico_loss}.

\begin{figure*}[!th]
     \centering
      \vspace{-5pt}
      \includegraphics[width=0.98\textwidth]{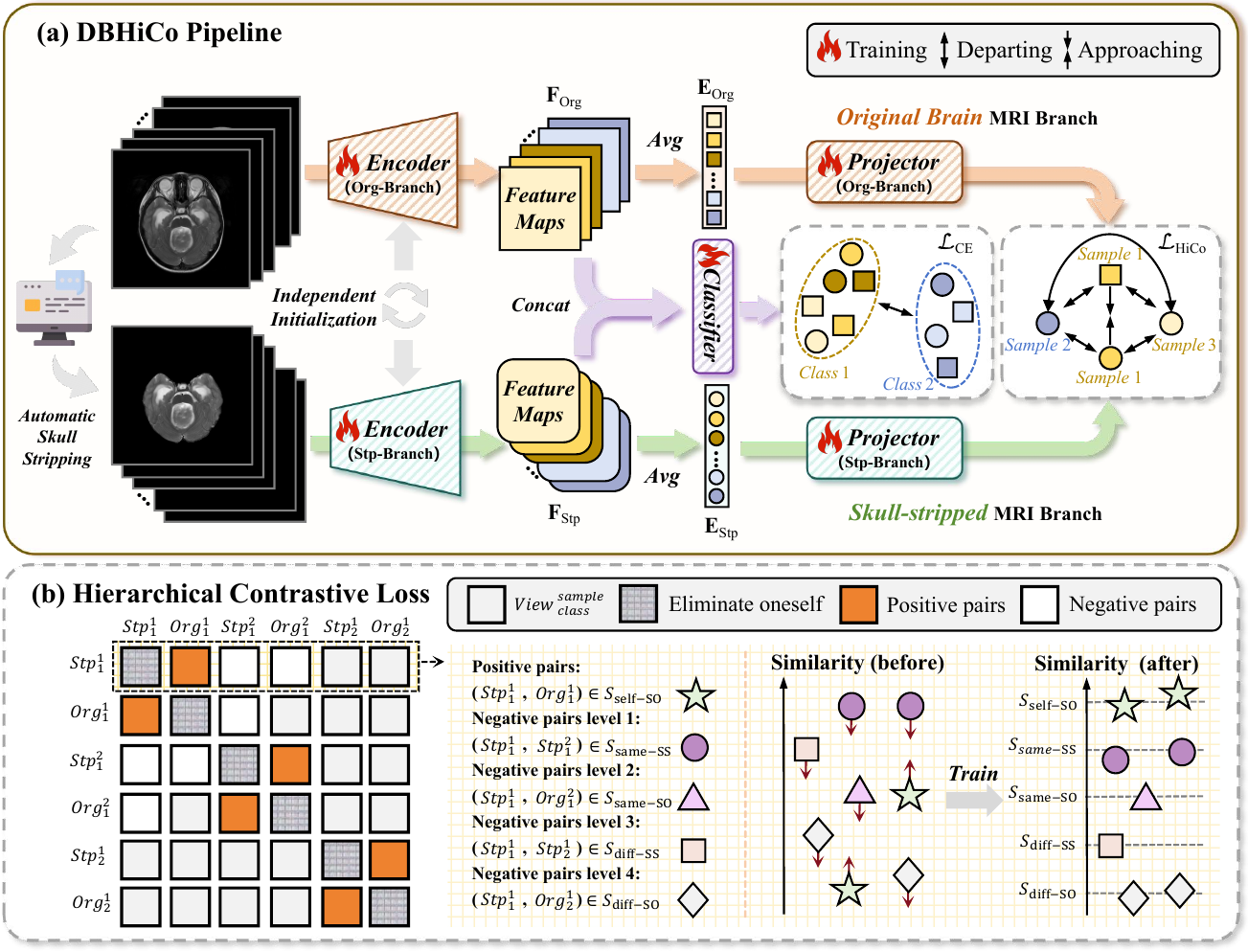}
      \vspace{-5pt}
      \caption{An overview of DBHiCo. \textbf{(a)} The pipeline of proposed DBHiCo. A dual-branch pipeline processes original and skull-stripped MRIs with independent encoders, extracting features for classification (\(\mathcal{L}_\text{CE}\)) and contrastive learning (\(\mathcal{L}_\text{HiCo}\)).
      \textbf{(b)} Hierarchical Contrastive Loss framework, which defines positive intra-subject pairs and multi-level negative pairs, enforcing structured similarity constraints to learn discriminative features.}
      \label{fig:Pipeline}
  \end{figure*}


\subsection{Overall Framework}\label{sec:framework}
The overall framework of the proposed DBHiCo method is presented in Figure~\hyperref[fig:Pipeline]{\ref*{fig:Pipeline}-a}. 
First, we establish a dual-branch encoder architecture to parallelly extract cross-view representations from the input brain MRI samples.
Specifically, let $\mathcal{D} = \left\{ \mathbf{I}_j^i \mid 1 \leq j \leq K, 1 \leq i \leq N_j \right\}$ denote a labeled MRI dataset which includes a set of $K$ class labels $y_{i} \in \{ 1,2,...,K\}$, with $N_j$ samples per class. We perform basic pre-processing steps (see Section~\ref{sec:datasets}) to obtain the $\text{view} \in \left\{ \text{Org}, \text{Stp} \right\}$ of original samples $\mathcal{O}= \left\{ \mathbf{I}_\text{Org} \right\}$ and corresponding skull-stripped samples $\mathcal{S} = \left\{ \mathbf{I}_{\text{Stp}} \right\}$. Each MRI image is represented as a 3D volume $\mathbf{I}_{\text{view}} = \left\{ \mathbf{x}_{\text{view,t}}  \right\}_{t=1}^{T}$, formed by stacking 2D slices $\mathbf{x} \in \mathbb{R}^{C \times H \times W}$ of $T$ frames. A typical encoder (\textit{e.g.}, CNN~\citep{he2016deep} or Transformer~\citep{dosovitskiy2020image}) is used to transform each single view image \textit{\textbf{frame-by-frame}} into a high-dimension feature map $\mathbf{F} = \{ \mathbf{f}_t \}_{t=1}^{T} \in \mathbb{R}^{T \times D}$. Second, on the one side, the cross-view representations are concatenated as $\mathbf{F}_{\text{comb}} = \left[ \mathbf{F}_{\text{Org}}; \mathbf{F}_{\text{Stp}} \right] \in \mathbb{R}^{T \times 2D}$ along the feature dimension $D$, after which a classifier generates \textbf{\textit{frame-level predictions}} $P = \frac{1}{T} \sum_{t=1}^{T} p_{t}$. On the other side, we perform \textit{\textbf{frame-averaging}} to obtain the contrast embedding $\mathbf{E}_{\text{view}} = \frac{1}{T} \sum_{t=1}^{T} \mathbf{F}_{\text{view},t} \in \mathbb{R}^{1 \times D} $, followed by the utilization of non-linear projectors (\textit{e.g.}, MLP) to map it into a lower-dimensional $\tilde{D}$ embedding $\mathbf{z}_{\text{view}} = \text{Proj}(\mathbf{E}_{\text{view}}) \in \mathbb{R}^{1 \times \tilde{D}}$. Finally, the objective function is formulated as:
\begin{equation}
\mathcal{L} = \mathcal{L}_{\text{CE}}(P, y) +  \alpha \cdot \mathcal{L}_{\text{HiCo}}(\mathbf{z}_{\text{Org}}, \mathbf{z}_{\text{Stp}}, y), 
\end{equation} where $\mathcal{L}_{\text{CE}}$ is the cross-entropy loss, and $\mathcal{L}_{\text{HiCo}}$ is our dedicated hierarchical contrast loss with weight $\alpha$. Further details of $\mathcal{L}_{\text{HiCo}}$ are presented in Section~\ref{sec:construct_pairs} and Section~\ref{sec:hico_loss}.

\subsection{Constructing Positive and Negative Pairs}\label{sec:construct_pairs}
The construction strategy of positive and negative pairs is crucial for effectively leveraging CL to extract discriminative and robust representations from brain MRI samples. The most common strategy~\citep{chen2020simple} is to treat different augmented views of the same image as a positive pair, while augmented views from different samples are negative pairs. To establish complementary cross-view relationships in brain MRI, we treat only the original and skull-stripped augmented views of the same image as positive pairs. Specifically, given the $i^{th}$ sample $(\textbf{I}_j^i, y_i)$ of the $j^{th}$ class in $\mathcal{D}_{\text{Train}}$, the unique positive pair is defined as $({\mathbf{I}_{\text{Org}}}_j^i, {\mathbf{I}_{\text{Stp}}}_j^i)$. The corresponding negative pairs are constructed as:
\begin{equation}
\begin{split}
\mathcal{N}_{i,j} = \bigl\{ & ({\mathbf{I}_{\text{Org}}}_j^i, {\mathbf{I}_{\text{Org}}}_m^n),\, ({\mathbf{I}_{\text{Org}}}_j^i, {\mathbf{I}_{\text{Stp}}}_m^n), \,
                           ({\mathbf{I}_{\text{Stp}}}_j^i,  {\mathbf{I}_{\text{Org}}}_m^n),\\&\, ({\mathbf{I}_{\text{Stp}}}_j^i, {\mathbf{I}_{\text{Stp}}}_m^n)  \mid (m, n) \neq (i, j) \bigr\},
\end{split}
\end{equation}
where \((m, n) \neq (i, j)\) ensures that negatives are drawn from different samples or classes. 
We assign difficulty levels to negative pairs according to three prior conditions:
\begin{enumerate}
  \item   \textbf{\textit{Cross-view consistency priority}}. Within each sample, the Org and Stp counterparts should remain consistent, ensuring that intra-sample cross-view similarity is preserved.
  \item \textbf{\textit{Intra-view robustness}}. For samples of the same class, the intra-view pairs should exhibit higher similarity than cross-view pairs, thereby reinforcing structural-consistency learning.
  \item \textbf{\textit{Discriminative constraint}}. Intra-class similarity should, overall, be higher than inter-class similarity, maintaining the class separability.
\end{enumerate}
This prior design enables us to rank the difficulties. Taking the Stp samples as an example (see Figure~\hyperref[fig:Pipeline]{\ref*{fig:Pipeline}-b}), the ordering of its associated positive and negative pairs can be illustrated as follows:  
\begin{equation}\label{order}
\begin{aligned} 
({\mathbf{I}_{\text{Stp}}}_j^i, {\mathbf{I}_{\text{Org}}}_j^i) &> \mathbb{E}_{\substack{i \neq n,  j = m}} ({\mathbf{I}_{\text{Stp}}}_j^i, {\mathbf{I}_{\text{Stp}}}_m^n) \\
&> \mathbb{E}_{\substack{i \neq n,  j = m}} ({\mathbf{I}_{\text{Stp}}}_j^i, {\mathbf{I}_{\text{Org}}}_m^n) \\
&> \mathbb{E}_{\substack{i \neq n , j \neq m}} ({\mathbf{I}_{\text{Stp}}}_j^i, {\mathbf{I}_{\text{Stp}}}_m^n) \\
&> \mathbb{E}_{\substack{i \neq n, j \neq m}}({\mathbf{I}_{\text{Stp}}}_j^i, {\mathbf{I}_{\text{Org}}}_m^n) \\
\end{aligned}
\end{equation} where $\mathbb{E}[\cdot]$ denotes the average over the specified image sets.

\subsection{Hierarchical Contrastive Loss}\label{sec:hico_loss}
\noindent\textbf{Preliminary: Contrastive Loss.}\label{sec:Preliminary}
Given two sets of projected contrast embedding $\left\{ \mathbf{z}_a^{(1)} \right\}_{a=1}^{N_B}  \in \mathbb{R}^{1 \times \tilde{D}} $  and $\left\{ \mathbf{z}_b^{(2)} \right\}_{b=1}^{N_B}  \in \mathbb{R}^{1 \times \tilde{D}}$ of the two augmented views in the batch ${N_B}$, the contrastive loss maximizes the similarity of positive pairs while minimizing that of negative pairs. Mathematically, the similarity can be calculated as:
\begin{equation}
sim\left(\mathbf{z}_a^{(1)}, \mathbf{z}_b^{(2)}\right) = \frac{\left(\mathbf{z}_a^{(1)}\right)^{\top} \mathbf{z}_b^{(2)}}{\tau},
\end{equation} where $sim(\cdot, \cdot)$ denotes the similarity function and $\tau$ is the temperature scaling factor that controls the concentration of the similarity distribution. Construct a row of logits $\ell_{a,b} = \left\{ sim\left(\mathbf{z}_a^{(1)}, \mathbf{z}_b^{(2)}\right)\right\}_{b=1}^{N_B}$ with the $a^{th}$ sample as the anchor, where the positive sample is $\ell_{a,a}$. Then, the standard contrastive (\textit{e.g.}, InfoNCE~\citep{oord2018representation}) loss is:
\begin{equation}
\mathcal{L}_{\text{Con}} = \frac{1}{N_B} \sum_{b=1}^{N_B} \left[ -\log \frac{\exp(\ell_{a,a})}{\sum_{b=1}^{N_B} \exp(\ell_{a,b})} \right].
\end{equation} which encourages the model to distinguish the positive sample (diagonal entry) from all others in the batch.

\noindent\textbf{Hierarchical Contrastive Loss.}
Building upon the basic contrastive loss, the proposed hierarchical contrastive loss $\mathcal{L}_{\text{HiCo}}$ introduces a structured similarity constraint to model fine-grained semantic hierarchies in cross-view MRI samples. 
Let $\mathbf{z}_{\text{Org}} \in \mathbb{R}^{{N_B} \times \tilde{D}}$ and $\mathbf{z}_{\text{Stp}} \in \mathbb{R}^{{N_B} \times \tilde{D}}$ denote the normalized feature embeddings of original and skull-stripped MRI views within a batch of size $N_B$, respectively, and let $\mathbf{l} \in \mathbb{R}^{N_B} $ be the corresponding class label vector, where $\mathbf{l}_a$ and $\mathbf{l}_b$ represent the labels of the $a^{th}$ and $b^{th}$ samples. Consider the $a^{th}$ image $\mathbf{z}_{\text{Stp}}$ as the anchor, the $\mathcal{L}_{\text{HiCo}}$ imposes a strict similarity ordering across five hierarchical levels:
\begin{enumerate}
    \item \textbf{\textit{Self cross-view similarity}}. 
    Similarity between the cross-view of the sample itself: 
    \begin{equation}
        S_{\text{self-SO}}^a =  sim(\mathbf{z}_{\text{Stp}}^a, \mathbf{z}_{\text{Org}}^a).
    \end{equation}

    \item \textbf{\textit{Intra-class and intra-view similarity}}. 
    Average similarity between the $\mathbf{z}_{\text{Stp}}$ of the $a^{th}$ sample and the others in the same class: 
    \begin{equation}
        S_{\text{same-SS}}^a = \frac{\sum_{b \neq a, \, \mathbf{l}_b = \mathbf{l}_a} sim(\mathbf{z}_{\text{Stp}}^a, \mathbf{z}_{\text{Stp}}^b)}{\text{count}(\{b \neq a, \, \mathbf{l}_b = \mathbf{l}_a\})}.
    \end{equation}
    
    \item \textbf{\textit{Intra-class and cross-view similarity}}. 
    Average similarity between the $\mathbf{z}_{\text{Stp}}$ of the $a^{th}$ sample and $\mathbf{z}_{\text{Org}}$ of the others in the same class: 
    \begin{equation}
        S_{\text{same-SO}}^a = \frac{\sum_{b \neq a, \, \mathbf{l}_b = \mathbf{l}_a} sim(\mathbf{z}_{\text{Stp}}^a, \mathbf{z}_{\text{Org}}^b)}{\text{count}(\{b \neq a, \, \mathbf{l}_b = \mathbf{l}_a\})}.
    \end{equation}

    \item \textbf{\textit{Inter-class and intra-view similarity}}. 
    Average similarity between the $\mathbf{z}_{\text{Stp}}$ of the $a^{th}$ sample and $\mathbf{z}_{\text{Stp}}$ of the $a^{th}$ of the others in different classes: 
    \begin{equation}    
        S_{\text{diff-SS}}^a = \frac{\sum_{b, \, \mathbf{l}_b \neq \mathbf{l}_a} sim(\mathbf{z}_{\text{Stp}}^a, \mathbf{z}_{\text{Stp}}^b)}{\text{count}(\{b, \, \mathbf{l}_b \neq \mathbf{l}_a\})}.
    \end{equation}

    \item \textbf{\textit{Inter-class and cross-view similarity}}. 
    Average similarity between the $\mathbf{z}_{\text{Stp}}$ of the $a^{th}$ sample and $\mathbf{z}_{\text{Org}}$ of the other in different classes: 
    \begin{equation} 
        S_{\text{diff-SO}}^a = \frac{\sum_{b, \, \mathbf{l}_b \neq \mathbf{l}_a} sim(\mathbf{z}_{\text{Stp}}^a, \mathbf{z}_{\text{Org}}^b)}{\text{count}(\{b, \, \mathbf{l}_b \neq \mathbf{l}_a\})}.
    \end{equation}
    
\end{enumerate}

The $\mathcal{L}_{\text{HiCo}}$ enforces a strict ordering according to \eqref{order} of these similarities with a margin $m$. Violations of these constraints are penalized using the $ReLU$ function. 
The single-directional loss is defined as the average of these penalty terms across all samples in the batch:
\begin{equation}
    \mathcal{L}_{\text{single}}(\mathbf{z}_{\text{Stp}}, \mathbf{z}_{\text{Org}}) = \frac{1}{N_B} \sum_{a=1}^{N_B} (\mathcal{L}_1 + \mathcal{L}_2 + \mathcal{L}_3 + \mathcal{L}_4)
\end{equation} 
where
\begin{equation}
\begin{aligned} 
    & \mathcal{L}_1 =  ReLU(-(S_{\text{self-SO}}^a-S_{\text{same-SS}}^a)+m) \cdot \mathbb{I}_{\text{same}}, \\
    & \mathcal{L}_2 =  ReLU(-(S_{\text{same-SS}}^a-S_{\text{same-SO}}^a)+m) \cdot \mathbb{I}_{\text{same}}, \\
    & \mathcal{L}_3 =  ReLU(-(S_{\text{same-SO}}^a-S_{\text{diff-SS}}^a)+m) \cdot \mathbb{I}_{\text{same}}, \\
    & \mathcal{L}_4 =  ReLU(-(S_{\text{diff-SS}}^a-S_{\text{diff-SO}}^a)+m) \cdot \mathbb{I}_{\text{diff}}. \\
\end{aligned}    
\end{equation} 
Here, $\mathbb{I}_{\text{same}}$ and $\mathbb{I}_{\text{diff}}$ are indicator functions that equal $1$ if there exist valid intra-class and inter-class samples for the $a^{th}$ sample, respectively, and $0$ otherwise. This ensures the margin loss is computed only when sufficient samples are available to form the respective pairs. Finally, the symmetric $\mathcal{L}_{\text{HiCo}}$ is expressed as:
\begin{equation}
    \mathcal{L}_{\text{HiCo}} = \frac{1}{2} \Big( \mathcal{L}_{\text{single}}(\mathbf{z}_{\text{Stp}}, \mathbf{z}_{\text{Org}}) + \mathcal{L}_{\text{single}}(\mathbf{z}_{\text{Org}}, \mathbf{z}_{\text{Stp}}) \Big).
\end{equation} 


\section{Experiments}\label{sec:experiments}

This section first introduces the datasets and evaluation metrics (Section~\ref{sec:datasets}), and then describes the implementation details of the proposed DBHiCo (Section~\ref{sec:implementation}). Finally, we report the comparative experimental results against existing methods (Section~\ref{sec:comparison}), followed by a comprehensive analysis of our approach (Section~\ref{sec:analysis}).

\begin{table*}[!t]
    \centering
    \caption{Experiment results on Tiantan dataset across four tumors. 
    All methods are trained on both original and skull-stripped MRI scans to ensure fairness.
    The best result is highlighted in \textbf{bold}, while the second-best is marked with an \underline{underline}.  
    }
    \label{tab:tiantan_tumor}
    \resizebox{1\linewidth}{!}{
        \begin{tabular}{lccccccccccccc}
            \whline
            \multirow{2}{*}{Method (Publications,Years)} & \multirow{2}{*}{\makecell[c]{Pre-training \\dataset}} & \multirow{2}{*}{Network} & \multirow{2}{*}{Input Size} & \multicolumn{2}{c}{Medulloblastoma} & \multicolumn{2}{c}{Choroid} & \multicolumn{2}{c}{Ependymoma} & \multicolumn{2}{c}{Glioma} & \multicolumn{2}{c}{Total} \\
            \cmidrule(r){5-6} \cmidrule(lr){7-8}  \cmidrule(lr){9-10}  \cmidrule(lr){11-12}  \cmidrule(lr){13-14}
            
            & & & & ACC(\%) & F1(\%) & ACC(\%) & F1(\%) & ACC(\%) & F1(\%) & ACC(\%) & F1(\%) & ACC(\%) & F1(\%) \\ 
            \hline
            \textit{\textbf{MRI Multi-View Representation Learning}} \\
            DAE (MIDL 2024)~\citep{valanarasu2023disruptive}        & DAE-10K~\citep{valanarasu2023disruptive}        & Swin-Unetr      & $8 \times 224 \times 224$   & $90.40$             & $89.04$             & $55.33$             & $63.01$             & $69.99$             & $67.14$             & $79.80$             & $79.96$             & $79.96$             & $74.79$             \\
            VoCo (CVPR 2024)~\citep{wu2024voco}                     & VoCo-1.6K~\citep{wu2024voco}                    & Swin-Unetr      & $8 \times 224 \times 224$   & $89.69$             & $87.67$             & $71.14$             & $72.91$             & $64.94$             & $68.61$             & $76.28$             & $74.93$             & $79.86$             & $76.03$             \\
            MiM (TMI 2025)~\citep{zhuang2025mim}                    & MiM-10K~\citep{zhuang2025mim}                   & Swin-ViT        & $8 \times 224 \times 224$   & $86.76$             & $86.07$             & $75.19$             & $71.38$             & $61.51$             & $65.35$             & $77.71$             & $77.31$             & $78.42$             & $75.03$             \\
            BrainMVP (CVPR 2025)~\citep{rui2025multi}               & BrainMVP-16K~\citep{rui2025multi}               & Uniformer-small & $16 \times 224 \times 224$  & $84.06$             & $81.16$             & $60.47$             & $66.03$             & $60.40$             & $63.14$             & $71.60$             & $70.48$             & $73.95$             & $70.20$             \\
            \hline
            \textit{\textbf{Baseline}} \\
            ViT (ICLR 2021)~\citep{dosovitskiy2020image}            & ILSVRC2012                                     & DeiT-Small      & $8 \times 224 \times 224$   & \underline{$90.93$} & $88.43$             & $71.28$             & $76.02$             & $74.44$             & $73.84$             & $74.95$             & $78.43$             & $81.88$             & $79.18$             \\
            ViT Recipe (MICCAI 2024)~\citep{kunanbayev2024training} & Brain-2K~\citep{kunanbayev2024training}         & DeiT-Base-3D    & $128 \times 128 \times 128$ & $89.48$             & $84.99$             & $67.23$             & $67.99$             & $55.38$             & $63.34$             & $75.64$             & $74.23$             & $77.11$             & $72.64$             \\
            \hline
            \textit{\textbf{With General Contrastive fine-tuning}} \\
            SimCLR (ICML 2020)~\citep{chen2020simple}               & ILSVRC2012                                     & DeiT-Small      & $8 \times 224 \times 224$   & $89.47$             & $89.43$             & $70.33$             & $77.22$             & \underline{$80.22$} & $78.00$             & $79.83$             & $77.95$             & $83.43$             & $80.65$             \\
            SuperCon (NIPS 2020)~\citep{khosla2020supervised}       & ILSVRC2012                                     & DeiT-Small      & $8 \times 224 \times 224$   & $90.44$             & $90.28$             & $74.19$             & $77.38$             & $73.96$             & $75.91$             & $81.94$             & $77.75$             & $83.51$             & $80.33$             \\
            BCL (CVPR 2022)~\citep{zhu2022balanced}                 & ILSVRC2012                                     & DeiT-Small      & $8 \times 224 \times 224$   & $88.47$             & \underline{$90.53$} & \underline{$82.14$} & \underline{$81.30$} & $79.04$             & $\mathbf{79.30}$    & \underline{$83.34$} & $78.54$             & $84.71$             & \underline{$82.42$} \\
            Proco (TPAMI 2024)~\citep{du2024probabilistic}          & ILSVRC2012                                     & DeiT-Small      & $8 \times 224 \times 224$   & $87.72$             & $89.04$             & $74.28$             & $77.81$             & $79.13$             & $75.37$             & $79.82$             & $79.07$             & $82.89$             & $80.33$             \\
            \hline
            \textit{\textbf{With Medical Contrastive fine-tuning}} \\
            ECL (MICCAI 2023)~\citep{zhang2023ecl}                  & ILSVRC2012                                     & DeiT-Small      & $8 \times 224 \times 224$   & $88.71$             & $87.20$             & $78.09$             & $78.32$             & $70.05$             & $73.03$             & $77.71$             & $78.13$             & $81.55$             & $79.16$             \\
            BPaCo (MICCAI 2024)~\citep{cai2024bpaco}                & ILSVRC2012                                     & DeiT-Small      & $8 \times 224 \times 224$   & $89.48$             & $85.63$             & $65.38$             & $70.61$             & $62.21$             & $65.46$             & $75.67$             & $76.09$             & $78.32$             & $74.45$             \\
            CoCa (KBS 2024)~\citep{sun2024expert}                   & ILSVRC2012                                     & DeiT-Small      & $8 \times 224 \times 224$   & $89.71$             & $90.10$             & $78.14$             & $80.19$             & $77.26$             & $76.28$             & $80.46$             & \underline{$79.58$} & $84.08$             & $81.54$             \\
            \hline
            
            \textbf{HiCo (ours)}                                   & ILSVRC2012                                     & DeiT-Small      & $8 \times 224 \times 224$   & $90.92$             & $\mathbf{90.68}$    & $73.23$             & $78.37$             & $\mathbf{81.89}$    & \underline{$78.87$} & $79.82$             & $\mathbf{80.74}$    & $\mathbf{84.89}$    & $82.17$             \\
            \textbf{DBHiCo (ours)}                                 & ILSVRC2012                                     & DeiT-Small      & $8 \times 224 \times 224$   & $\mathbf{91.92}$    & $\mathbf{92.22}$    & $\mathbf{82.19}$    & $\mathbf{83.23}$    & $77.93$             & \underline{$78.08$} & $\mathbf{84.68}$    & $\mathbf{83.29}$    & $\mathbf{86.51}$    & $\mathbf{84.21}$    \\
            \whline
        \end{tabular}
    }
\end{table*}
\begin{figure*}[!th]
     \centering
      \includegraphics[width=\textwidth]{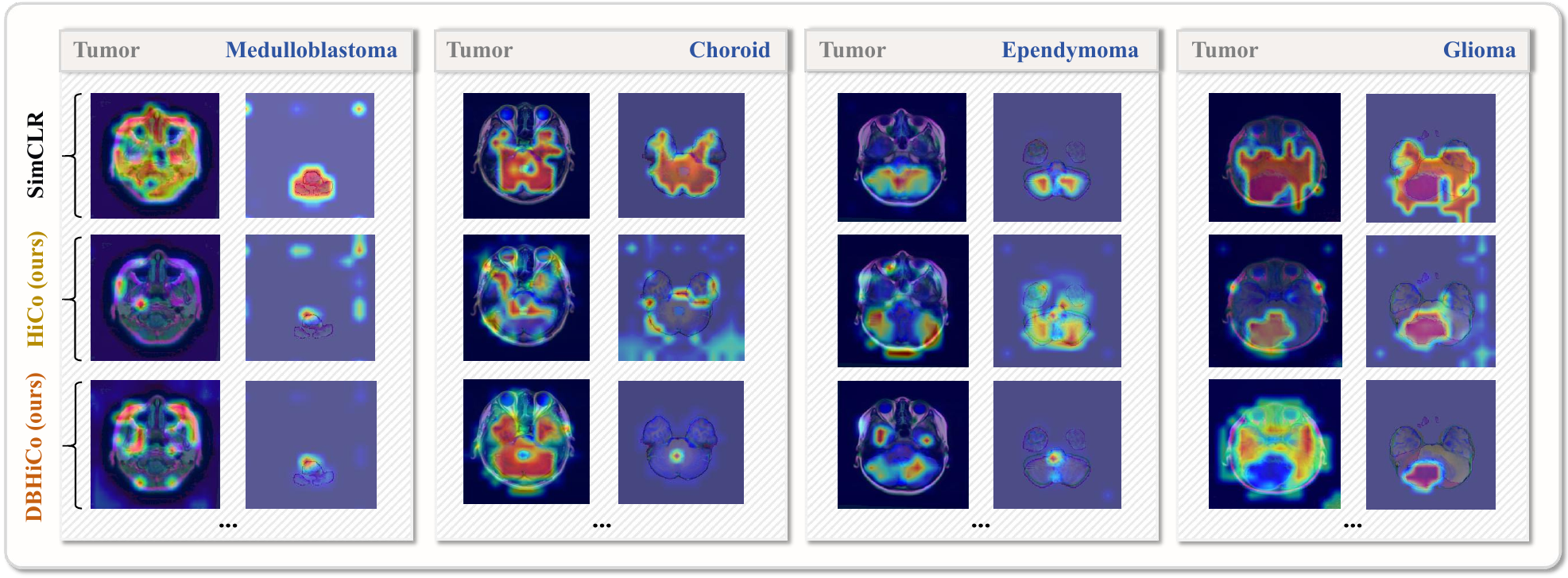}
      \vspace{-15pt}
      \caption{Grad-CAM~\citep{selvaraju2017grad} visualizations across four tumor types on the Tiantan dataset. While the baseline SimCLR is prone to background interference and exhibits blurred boundaries, our proposed HiCo effectively captures fine-grained structural details. Furthermore, DBHiCo achieves the most precise localization of anatomical structures and pathological tissues.}
      \label{fig:vis_attns}
  \end{figure*}

\subsection{Datasets and Evaluation}\label{sec:datasets}

\noindent\textbf{Dataset Collection and Pre-processing.}
In this study, we evaluate our method on three distinct types of brain disorders: Pediatric Posterior Fossa Tumors, Alzheimer’s Disease, and Attention-Deficit Hyperactivity Disorder. 

Due to the scarcity of publicly available datasets for Pediatric Posterior Fossa Tumor classification~\citep{jeong2019machine,nakamoto2019prediction,ismael2020enhanced,quon2020deep}, we compiled a proprietary clinical dataset, referred to as the \textbf{\textit{Tiantan}} dataset, sourced from the Department of Neurosurgery at Beijing Tiantan Hospital. This dataset comprises $407$ cases of medulloblastoma, $101$ cases of choroid plexus tumors, $177$ cases of ependymoma, and $144$ cases of glioma. 
This retrospective study was approved by the Institutional Review Board and the ethics committee of Beijing Tiantan Hospital, Capital Medical University (No. KY2022-067-01). 
Between December $2004$ and September $2021$, image acquisition was performed using three MRI scanners: Siemens Magnetom Verio, Siemens Magnetom TrioTim, and GE Discovery MR750. We ensured that each subject in the Tiantan dataset possessed three structural MRI modalities: T1-weighted (T1), contrast-enhanced T1-weighted (T1-E), and T2-weighted (T2) sequences. 
The acquisition parameters for the T1 sequence were: repetition time (TR) = $1900\sim2400$ ms, echo time (TE) = $8.6\sim19.84$ ms, slice thickness = $5$ mm, and matrix sizes of $512\times496$ (Magnetom Verio), $512\times432$ (Magnetom TrioTim), or $512\times512$ (Discovery MR750). 
The parameters for T1-E were: TR = $1900\sim2400$ ms, TE = $8.6\sim19.84$ ms, slice thickness = $5$ mm, with identical matrix sizes to T1. 
For the T2 sequence, parameters were: TR = $4500\sim10738.38$ ms, TE = $8.6\sim113.47$ ms, slice thickness = $5$ mm, and matrix sizes of $384\times324$ (Magnetom Verio), $640\times640$ (Magnetom TrioTim), or $512\times512$ (Discovery MR750). Slice gaps were $1$ mm on the Magnetom Verio and Discovery MR750, and $1.5$ mm on the Magnetom TrioTim. Gd-DTPA (dose: $0.1$ mmol/kg) was administered as the contrast agent for T1-E imaging.

For Alzheimer’s Disease classification, we utilized the widely adopted public dataset \textbf{\textit{OASIS-1}}~\citep{marcus2007open}. It contains $416$ subjects aged $18$ to $96$ years, including $100$ clinically diagnosed with Alzheimer's Disease. Each subject's data includes $3$ or $4$ T1-weighted scans acquired during a single imaging session. 
For Attention-Deficit Hyperactivity Disorder, we evaluated our method on the public \textbf{\textit{ADHD-200}}~\citep{adhd2012adhd} global competition dataset, originally aggregated from $8$ independent imaging sites. We included $691$ subjects with T1-weighted MRI scans and definitive diagnostic labels from $6$ of these sites, of which $261$ subjects were clinically diagnosed with ADHD.

Across all three datasets, our study focuses exclusively on cross-sectional structural MRI. 
For the Tiantan dataset, we initially applied the N4ITK~\citep{song2017review} bias field correction to mitigate intensity inhomogeneities and standardize intensity distributions across similar tissue types. Subsequently, a unified skull-stripping procedure was applied to all subjects across the three datasets. Since skull stripping crucially impacts the downstream accuracy of brain tumor and lesion recognition, we selected the robust HD-BET~\citep{isensee2019automated} algorithm over other traditional methods~\citep{smith2002fast,shattuck2001magnetic,eskildsen2012beast,roy2017robust} due to its superior generalization capabilities across heterogeneous scanners and varying MRI modalities.

\noindent\textbf{Evaluation metrics.}
Each dataset was partitioned into training, validation, and test sets. To ensure a robust evaluation, we performed five-fold cross-validation and reported the average test-set performance across all folds for each compared method. While Accuracy (ACC) and the Area Under the Receiver Operating Characteristic Curve (AUC) are standard metrics for disease classification, medical datasets often suffer from severe class imbalance. To address the inherent biases of ACC and AUC under such distributions, we additionally incorporated the macro-averaged F1-score (F1) as a comprehensive evaluation metric.

\begin{table*}[!th]
    \centering
    \caption{Experiment results on OASIS-I dataset~\citep{marcus2007open}. }
    \label{tab:OASIS}
    \resizebox{1\linewidth}{!}{
        \begin{tabular}{lcccccccccc}
            \whline
            \multirow{2}{*}{Method (Publications,Years)}           & \multirow{2}{*}{\makecell[c]{Pre-training \\dataset}} & \multirow{2}{*}{Network} & \multirow{2}{*}{Input Size} & \multicolumn{2}{c}{Non Demented} & \multicolumn{2}{c}{Demented} & \multicolumn{3}{c}{Total} \\
            \cmidrule(r){5-6} \cmidrule(lr){7-8}  \cmidrule(lr){9-11}
            
            & & & & ACC(\%) & F1(\%) & ACC(\%) & F1(\%) & ACC(\%) & F1(\%) & AUC(\%) \\[-1pt]
            \hline
            \textit{\textbf{MRI Multi-View Representation Learning}} \\
            DAE (MIDL 2024)~\citep{valanarasu2023disruptive}       & DAE-10K~\citep{valanarasu2023disruptive}              & Swin-Unetr               & $96 \times 96 \times 96$    & $82.91$             & $86.93$             & $75.00$             & $65.21$             & $81.02$             & $76.07$             & $87.42$             \\
            VoCo (CVPR 2024)~\citep{wu2024voco}                    & VoCo-1.6K~\citep{wu2024voco}                          & Swin-Unetr               & $96 \times 96 \times 96$    & $82.28$             & $86.23$             & $74.00$             & $64.14$             & $80.48$             & $75.19$             & $87.64$             \\
            MiM (TMI 2025)~\citep{zhuang2025mim}                   & MiM-10K~\citep{zhuang2025mim}                         & Swin-ViT                 & $8 \times 224 \times 224$   & $80.37$             & $84.64$             & $71.00$             & $60.94$             & $78.12$             & $72.79$             & $84.21$             \\
            BrainMVP (CVPR 2025)~\citep{rui2025multi}              & BrainMVP-16K~\citep{rui2025multi}                     & Uniformer-small          & $96 \times 96 \times 96$    & $76.56$             & $82.95$             & $76.99$             & $61.45$             & $76.68$             & $72.21$             & $84.66$             \\
            \hline
            \textit{\textbf{Baseline}} \\
            ResNet (CVPR 2016)~\citep{he2016deep}                  & ILSVRC2012                                            & ResNet-50                & $8 \times 224 \times 224$   & $83.54$             & $86.78$             & $72.99$             & $64.76$             & $81.11$             & $75.78$             & $86.45$             \\
            ViT (ICLR 2021)~\citep{dosovitskiy2020image}           & ILSVRC2012                                            & DeiT-Small               & $8 \times 224 \times 224$   & $87.34$             & $87.31$             & $60.00$             & $60.15$             & $80.56$             & $73.67$             & $85.30$             \\  
            ViT Recipe (MICCAI 2024)~\citep{kunanbayev2024training}& Brain-2K~\citep{kunanbayev2024training}               & DeiT-Base-3D             & $128 \times 128 \times 128$ & $89.25$             & $87.11$             & $51.00$             & $54.79$             & $80.06$             & $70.96$             & $84.00$             \\
            \hline
            \textit{\textbf{With General Contrastive fine-tuning}} \\
            SimCLR (ICML 2020)~\citep{chen2020simple}              & ILSVRC2012                                            & ResNet-50                & $8 \times 224 \times 224$   & $86.38$             & $87.86$             & $66.99$             & $62.05$             & $81.66$             & $74.96$             & $84.93$             \\
            SuperCon (NIPS 2020)~\citep{khosla2020supervised}      & ILSVRC2012                                            & ResNet-50                & $8 \times 224 \times 224$   & $81.01$             & $85.96$             & \underline{$77.00$} & $65.28$             & $79.99$             & $75.62$             & $87.76$             \\
            BCL (CVPR 2022)~\citep{zhu2022balanced}                & ILSVRC2012                                            & ResNet-50                & $8 \times 224 \times 224$   & $79.13$             & $85.04$             & $\mathbf{80.00}$    & $65.46$             & $79.36$             & $75.25$             & $86.70$             \\
            Proco (TPAMI 2024)~\citep{du2024probabilistic}         & ILSVRC2012                                            & ResNet-50                & $8 \times 224 \times 224$   & $83.53$             & $87.42$             & $76.00$             & $66.22$             & $81.82$             & $76.82$             & \underline{$88.84$} \\
            \hline
            \textit{\textbf{With Medical Contrastive fine-tuning}} \\
            ECL (MICCAI 2023)~\citep{zhang2023ecl}                 & ILSVRC2012                                            & ResNet-50                & $8 \times 224 \times 224$   & $79.41$             & $85.41$             & $\mathbf{80.00}$    & $65.20$             & $79.56$             & $75.31$             & $88.29$             \\
            BPaCo (MICCAI 2024)~\citep{cai2024bpaco}               & ILSVRC2012                                            & ResNet-50                & $8 \times 224 \times 224$   & $83.54$             & $87.35$             & \underline{$77.00$} & \underline{$67.81$} & $81.97$             & \underline{$77.58$} & $88.34$             \\
            CoCa (KBS 2024)~\citep{sun2024expert}                  & ILSVRC2012                                            & ResNet-50                & $8 \times 224 \times 224$   & $\mathbf{94.01}$    & \underline{$89.05$} & $45.99$             & $55.25$             & \underline{$82.46$} & $72.15$             & $88.49$             \\
            \hline
            \textbf{HiCo (ours)}                                   & ILSVRC2012                                            & ResNet-50                & $8 \times 224 \times 224$   & $85.75$             & $88.54$             & $74.99$             & $67.70$             & $\mathbf{83.17}$    & $\mathbf{78.12}$    & $\mathbf{90.21}$    \\
            \textbf{DBHiCo (ours)}                                 & ILSVRC2012                                            & ResNet-50                & $8 \times 224 \times 224$   & \underline{$89.88$} & $\mathbf{90.49}$    & $72.00$             & $\mathbf{69.70}$    & $\mathbf{85.55}$    & $\mathbf{80.10}$    & $\mathbf{91.13}$    \\
            \whline
        \end{tabular}
    }
\end{table*}

\subsection{Implementation Details}\label{sec:implementation}
As outlined in Section~\ref{sec:framework}, DBHiCo employs a standard backbone network as the feature encoder, initialized with pre-trained weights from large-scale natural image datasets (\textit{e.g.}, ILSVRC-2012~\citep{russakovsky2015imagenet}). Originally, each MRI scan is a single-channel grayscale 3D volume (multi-slice image). To seamlessly adapt these inputs for the 2D pre-trained encoder, we constructed pseudo-color 3-channel inputs based on the dataset: for the Tiantan dataset, the three distinct MRI modalities (T1, CET1, T2) of each subject were stacked along the channel dimension; for the OASIS-1 dataset, three separate T1-weighted scans from the same session were randomly selected and stacked; and for the ADHD-200 dataset, the single T1-weighted scan was replicated across all three channels. By default, we uniformly sampled $8$ slices along the z-axis from each subject. This strategy guarantees adequate spatial coverage of the entire brain volume while minimizing computational overhead and data redundancy.

\noindent\textbf{Hyperparameter setting.} 
For network optimization, the model was trained using the AdamW optimizer with a batch size of $8$, an initial learning rate of $0.001$, and a weight decay of $0.05$, regulated by a cosine annealing learning rate scheduler. The contrastive projector was configured as an MLP comprising two linear layers separated by a ReLU activation function. Its intermediate hidden dimension was set to $512$, and the final output embedding dimension $\tilde{D}$ was set to $128$. For the proposed hierarchical contrastive loss ($\mathcal{L}_{\text{HiCo}}$), the temperature scaling factor $\tau$ was fixed at $1.0$, and the margin $m$ was empirically set to $0.1$.

\noindent\textbf{Comparison methods.}  
We rigorously compared our DBHiCo with several established baselines, state-of-the-art specialized MRI classification networks, and advanced contrastive learning frameworks. 
First, we established standard supervised baselines using ViT~\citep{dosovitskiy2020image} and ResNet~\citep{he2016deep}, alongside a ViT Recipe~\citep{kunanbayev2024training} specifically tailored for the medical domain. 
Second, we compared our approach with state-of-the-art MRI multi-view representation learning methods—including DAE~\citep{valanarasu2023disruptive}, VoCo~\citep{wu2024voco}, MiM~\citep{zhuang2025mim}, and BrainMVP~\citep{rui2025multi}—as these also explicitly model cross-view consistency in MRI. For these specialized methods, we preserved the backbone architectures specified in their respective publications (\textit{e.g.}, Swin-UNETR~\citep{hatamizadeh2021swin}, Swin-ViT~\citep{liu2021swin}, and Uniformer~\citep{li2022uniformer}), initializing them with pre-trained weights derived from the large-scale MRI/CT datasets constructed in their original studies. 
Third, we evaluated advanced contrastive learning paradigms, including SimCLR~\citep{chen2020simple} and SupCon~\citep{khosla2020supervised}, as well as BCL~\citep{zhu2022balanced} and Proco~\citep{du2024probabilistic}, which are specifically engineered to handle class imbalance. 
Finally, we included state-of-the-art medical contrastive approaches such as ECL~\citep{zhang2023ecl}, BPaCo~\citep{cai2024bpaco}, and CoCa~\citep{sun2024expert}. 
Although several of these contrastive methods conceptually describe a dual-branch setup, they practically implement weight-sharing encoders; hence, we categorized them as single-branch frameworks to maintain comparative clarity. 
For all evaluated contrastive methods, we standardized the backbone using DeiT-small~\citep{touvron2021training} or ResNet-50~\citep{he2016deep} initialized with ILSVRC-2012 weights. Importantly, to ensure a \textbf{strictly fair comparison}, all competing methods were trained utilizing both the original and the skull-stripped MRI scans.

\begin{table*}[htb]
    \centering
    \caption{Experiment results on ADHD200 dataset~\citep{adhd2012adhd} across six independent imaging sites.}
    \label{tab:ADHD200}
    \resizebox{1\textwidth}{!}{
        \begin{tabular}{lcccccccccccc}
            \whline
            \multirow{2}{*}{Method (Publications,Years)}           & \multirow{2}{*}{\makecell[c]{Pre-traning \\dataset}} & \multirow{2}{*}{Network} & \multirow{2}{*}{Input Size} & \multicolumn{6}{c}{ACC(\%)} & \multicolumn{3}{c}{Avg} \\
            \cmidrule(r){5-10} \cmidrule(lr){11-13}
            & & & & KKI & Neuro & NYU & OHSU & Peking & Pitts & ACC(\%) & F1(\%) & AUC(\%) \\ 
            \hline
            \textit{\textbf{MRI Multi-View Representation Learning}} \\
            DAE (MIDL 2024)~\citep{valanarasu2023disruptive}       & DAE-10K~\citep{valanarasu2023disruptive}             & Swin-Unetr               & $96 \times 96 \times 96$    & $\mathbf{74.57}$    & $63.51$             & $56.40$             & $62.47$             & $59.05$             & \underline{$100.0$} & $66.33$             & $63.00$             & $71.32$             \\
            VoCo (CVPR 2024)~\citep{wu2024voco}                    & VoCo-1.6K~\citep{wu2024voco}                         & Swin-Unetr               & $96 \times 96 \times 96$    & $68.99$             & \underline{$71.51$} & $59.08$             & $61.99$             & $64.92$             & \underline{$100.0$} & \underline{$67.91$} & $64.93$             & $71.42$             \\
            MiM (TMI 2025)~\citep{zhuang2025mim}                   & MiM-10K~\citep{zhuang2025mim}                        & Swin-ViT                 & $8 \times 224 \times 224$   & $66.90$             & $58.83$             & $53.64$             & $58.19$             & $\mathbf{68.64}$    & $98.88$             & $65.12$             & $62.37$             & $69.21$             \\
            BrainMVP (CVPR 2025)~\citep{rui2025multi}              & BrainMVP-16K~\citep{rui2025multi}                    & Uniformer-small          & $96 \times 96 \times 96$    & \underline{$73.32$} & $45.29$             & $60.96$             & $59.95$             & $61.13$             & \underline{$100.0$} & $66.43$             & $60.16$             & $69.68$             \\
            \hline
            \textit{\textbf{Baseline}} \\
            ViT (ICLR 2021)~\citep{dosovitskiy2020image}           & ILSVRC2012                                           & DeiT-Small               & $8 \times 224 \times 224$   & $68.81$             & $56.91$             & $55.39$             & $62.37$             & $65.41$             & \underline{$100.0$} & $65.97$             & $62.03$             & $72.21$             \\
            ViT Recipe (MICCAI 2024)~\citep{kunanbayev2024training}& Brain-2K~\citep{kunanbayev2024training}              & DeiT-Base-3D             & $128 \times 128 \times 128$ & \underline{$73.32$} & $40.51$             & $59.54$             & $58.62$             & $62.60$             & \underline{$100.0$} & $65.84$             & $61.04$             & $70.33$             \\
            \hline
            \textit{\textbf{With General Contrastive fine-tuning}} \\
            SimCLR (ICML 2020)~\citep{chen2020simple}              & ILSVRC2012                                           & DeiT-Small               & $8 \times 224 \times 224$   & $73.24$             & $57.55$             & $57.50$             & $58.72$             & $61.33$             & \underline{$100.0$} & $65.90$             & $64.63$             & $72.59$             \\
            SuperCon (NIPS 2020)~\citep{khosla2020supervised}      & ILSVRC2012                                           & DeiT-Small               & $8 \times 224 \times 224$   & $72.07$             & $63.09$             & $58.11$             & $63.68$             & $59.71$             & \underline{$100.0$} & $66.72$             & $64.81$             & $72.35$             \\
            BCL (CVPR 2022)~\citep{zhu2022balanced}                & ILSVRC2012                                           & DeiT-Small               & $8 \times 224 \times 224$   & $66.91$             & $64.05$             & $56.97$             & \underline{$65.28$} & \underline{$66.96$} & \underline{$100.0$} & $67.34$             & \underline{$65.66$} & $71.85$             \\
            Proco (TPAMI 2024)~\citep{du2024probabilistic}         & ILSVRC2012                                           & DeiT-Small               & $8 \times 224 \times 224$   & $72.01$             & $53.87$             & $54.27$             & $61.68$             & $63.31$             & \underline{$100.0$} & $65.03$             & $61.48$             & $65.96$             \\
            \hline
            \textit{\textbf{With Medical Contrastive fine-tuning}} \\
            ECL (MICCAI 2023)~\citep{zhang2023ecl}                 & ILSVRC2012                                           & DeiT-Small               & $8 \times 224 \times 224$   & $70.82$             & $58.05$             & $59.02$             & $59.50$             & $61.98$             & \underline{$100.0$} & $66.14$             & $64.07$             & $70.05$             \\
            BPaCo (MICCAI 2024)~\citep{cai2024bpaco}               & ILSVRC2012                                           & DeiT-Small               & $8 \times 224 \times 224$   & $62.28$             & $47.37$             & $56.83$             & $60.26$             & $54.68$             & \underline{$100.0$} & $62.53$             & $61.46$             & $70.00$             \\
            CoCa (KBS 2024)~\citep{sun2024expert}                  & ILSVRC2012                                           & DeiT-Small               & $8 \times 224 \times 224$   & $72.26$             & $62.72$             & $\mathbf{62.19}$    & $59.92$             & $59.62$             & \underline{$100.0$} & $67.01$             & $64.60$             & \underline{$73.40$} \\
             \hline
            \textbf{HiCo (ours)}                                   & ILSVRC2012                                           & DeiT-Small               & $8 \times 224 \times 224$   & $68.61$             & $\mathbf{80.18}$    & $58.27$             & $\mathbf{68.23}$    & $62.55$             & $\mathbf{100.0}$    & $\mathbf{68.73}$    & $\mathbf{66.93}$    & $\mathbf{73.86}$    \\
            \textbf{DBHiCo (ours)}                                 & ILSVRC2012                                           & DeiT-Small               & $8 \times 224 \times 224$   & \underline{$73.32$} & $\mathbf{78.73}$    & \underline{$60.73$} & $64.85$             & $63.19$             & $\mathbf{100.0}$    & $\mathbf{69.81}$    & $\mathbf{66.78}$    & $\mathbf{74.43}$    \\
            \whline
        \end{tabular}
    }\\
    \textsuperscript{*}{\ud{\raggedleft{\scriptsize{Note: KKI: Kennedy Krieger Institute, Neuro: NeuroIMAGE, NYU: New York University, OHSU: Oregon Health \& Science University, Peking: Peking University, Pitts: University of Pittsburgh.}}}}
\end{table*}

\subsection{Experiments on downstream tasks}\label{sec:comparison}
\subsubsection{Comparison on the Tiantan dataset}
We first conduct comparative experiments on the Tiantan dataset, with the quantitative results summarized in Table~\ref{tab:tiantan_tumor}. Notably, we observe that existing multi-view representation learning methods yield sub-optimal performance. For instance, the highest total ACC among them, achieved by DAE~\citep{valanarasu2023disruptive}, is even lower than that of the standard DeiT-Small~\citep{touvron2021training} baseline ($79.96\%$ vs. $81.88\%$). We attribute this underperformance to the discrepancy between their self-supervised pre-training tasks and the specific objectives of the downstream classification. Furthermore, their pre-training data (typically around 10K scans) are relatively limited compared to the million-scale ILSVRC-2012 dataset. Similarly, the heavier 3D ViT Recipe~\citep{kunanbayev2024training}, despite its more complex network architecture and larger parameter count, is prone to overfitting on our dataset, achieving only a $77.11\%$ total ACC.

When evaluating the re-implemented general and medical contrastive learning (CL) frameworks, BCL~\citep{zhu2022balanced} and CoCa~\citep{sun2024expert} emerge as the most competitive baselines, achieving the highest ($84.71\%$) and second-highest ($84.08\%$) total ACC, respectively. To ensure a fair structural comparison, we first evaluate a \textbf{shared-encoder variant} of our method, denoted as \textbf{HiCo}. Despite utilizing the same single-branch architecture as the baseline CL methods, HiCo achieves a superior $84.89\%$ total ACC. 
Furthermore, our full dual-branch architecture, \textbf{DBHiCo}, significantly outperforms all competitors, surpassing BCL and CoCa by $1.80\%$ and $2.43\%$ in total ACC, respectively. As evident in Table~\ref{tab:tiantan_tumor}, both HiCo and DBHiCo consistently deliver substantial improvements across nearly all per-class and overall metrics (ACC and F1-score). These strong results demonstrate that by jointly enforcing fine-grained structural and semantic constraints, our method enables a much more discriminative and robust understanding of MRI representations.

\subsubsection{Comparison on the OASIS-I datasets}

We further extend our evaluation to the OASIS-1~\citep{marcus2007open} dataset. As shown in Table~\ref{tab:OASIS}, existing MRI multi-view representation learning methods plateau at a maximum of $81.02\%$ ACC, $76.07\%$ F1, and $87.64\%$ AUC. Among the baseline models, we observe that ResNet-50~\citep{he2016deep} consistently outperforms the ViT-based architectures across all three overall metrics. This superiority can likely be attributed to the limited scale of this downstream dataset, where CNNs typically exhibit stronger inductive biases \citep{d2021convit} than Vision Transformers, making them less prone to overfitting. Consequently, we designate ResNet-50 as the standard backbone network for all subsequent contrastive learning comparisons on this dataset.

Notably, our proposed DBHiCo establishes a new state-of-the-art on this benchmark, surpassing all competing methods with an impressive $85.55\%$ ACC, $80.10\%$ F1, and $91.13\%$ AUC. Furthermore, even our single-branch HiCo variant demonstrates significant superiority, yielding absolute improvements of at least $0.71\%$ in ACC (against CoCa), $0.54\%$ in F1 (against BPaCo), and $1.37\%$ in AUC (against Proco) over the strongest baseline methods. Crucially, these striking improvements are attained without relying on any specialized mechanisms for handling class imbalance (unlike methods such as BCL or Proco), thereby underscoring the inherent robustness, scalability, and broad applicability of our hierarchical contrastive framework.



\begin{figure*}[!t]
\centering
\begin{minipage}[t]{0.24\textwidth}
\centering
\vspace{-3pt}
\includegraphics[width=\linewidth]{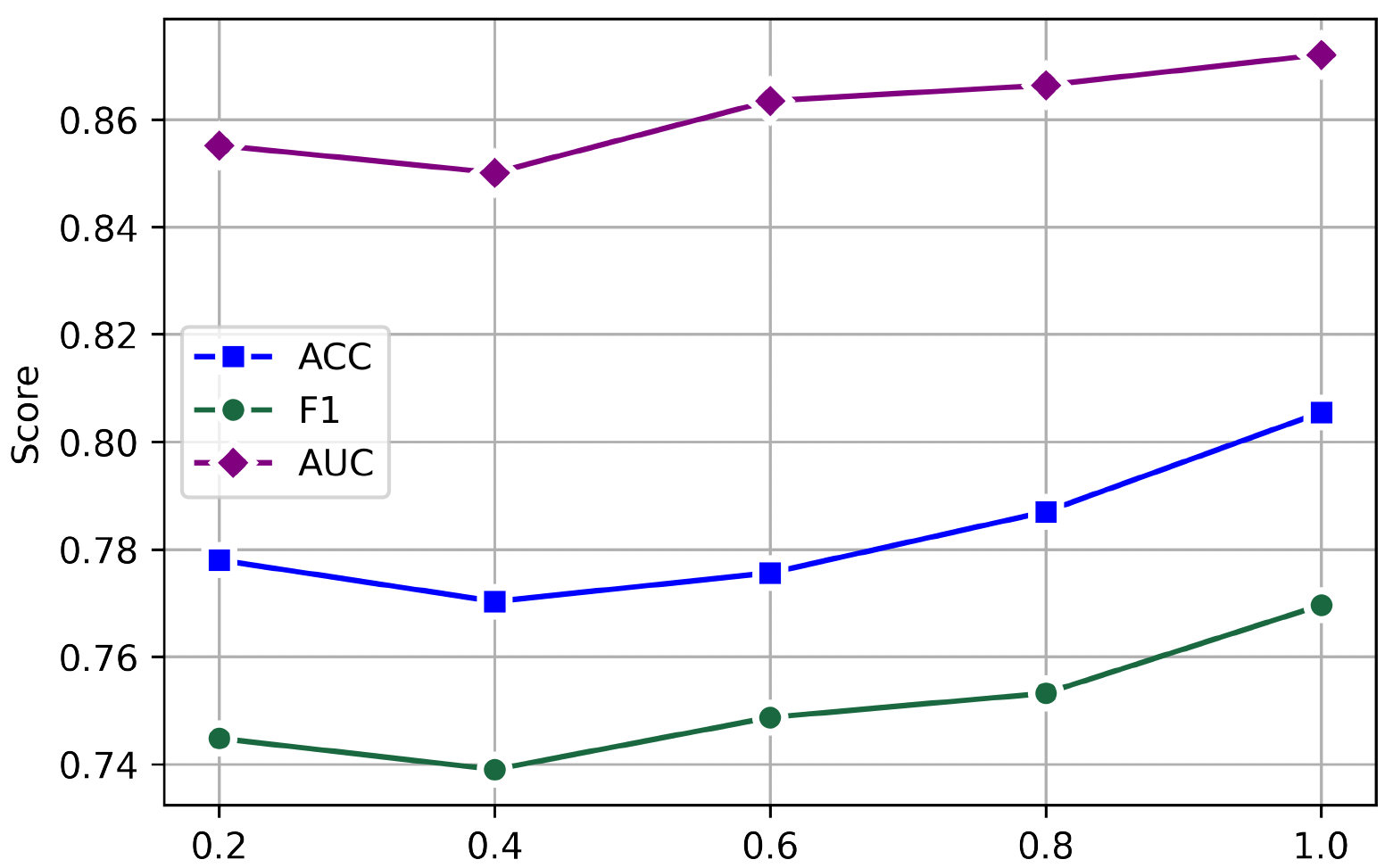}
\vspace{-3pt}
\small (a) AVG
\end{minipage}
\begin{minipage}[t]{0.24\textwidth}
\centering
\vspace{-3pt}
\includegraphics[width=\linewidth]{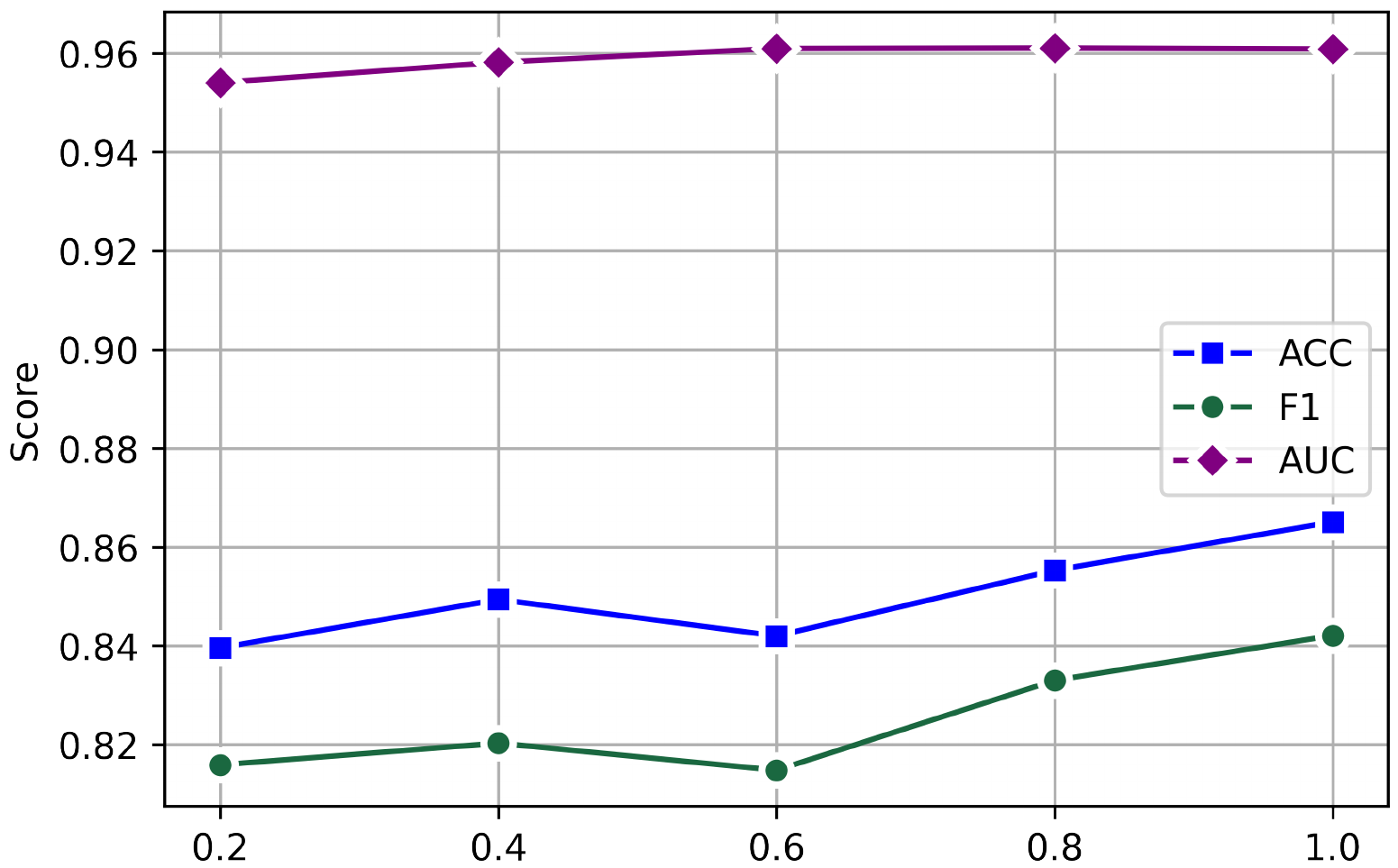}
\vspace{-3pt}
\small (b) Tiantan-Tumor Dataset
\end{minipage}%
\begin{minipage}[t]{0.24\textwidth}
\centering
\vspace{-3pt}
\includegraphics[width=\linewidth]{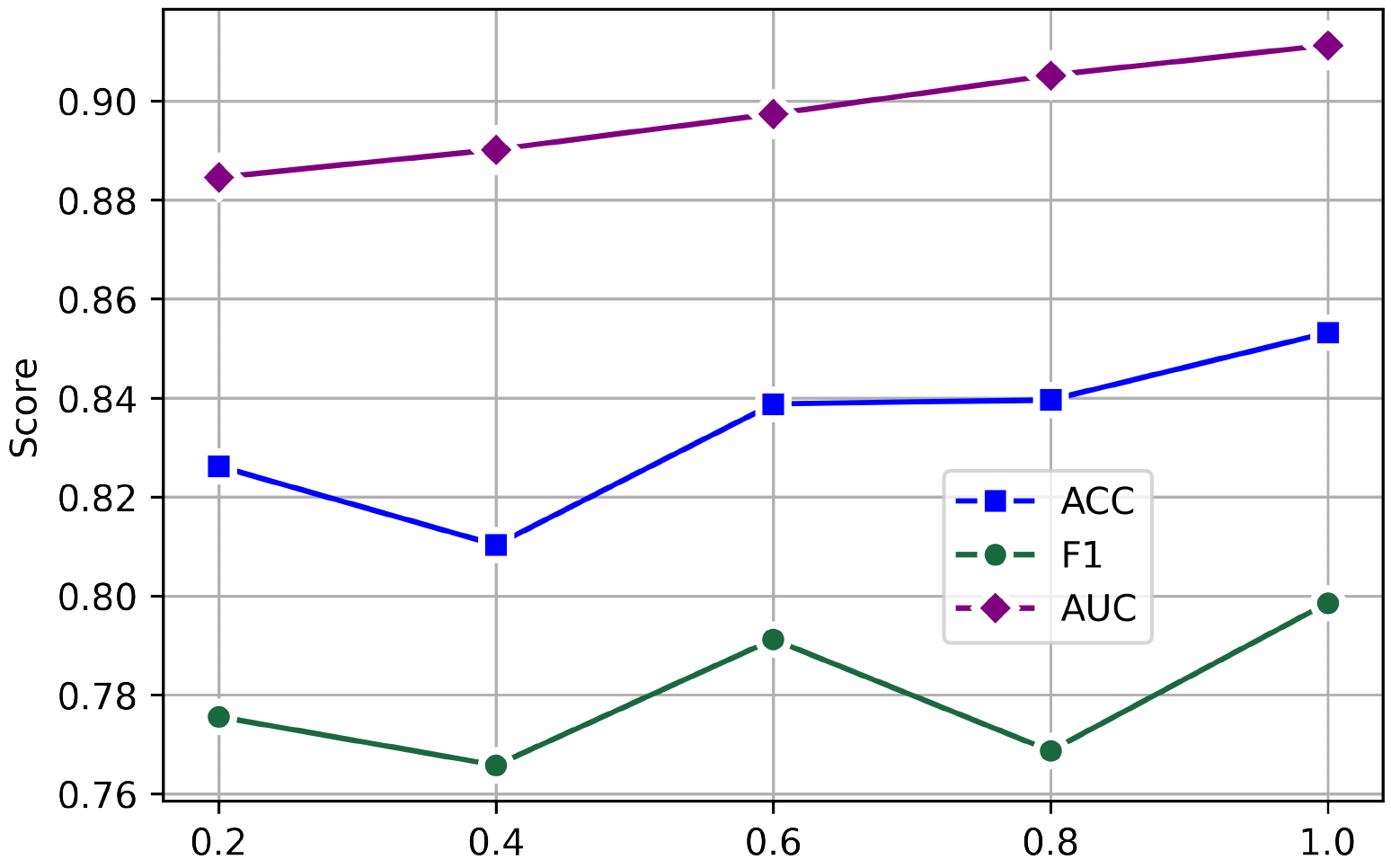}
\vspace{-3pt}
\small (c) OASIS-I Dataset
\end{minipage}%
\begin{minipage}[t]{0.24\textwidth}
\centering
\vspace{-3pt}
\includegraphics[width=\linewidth]{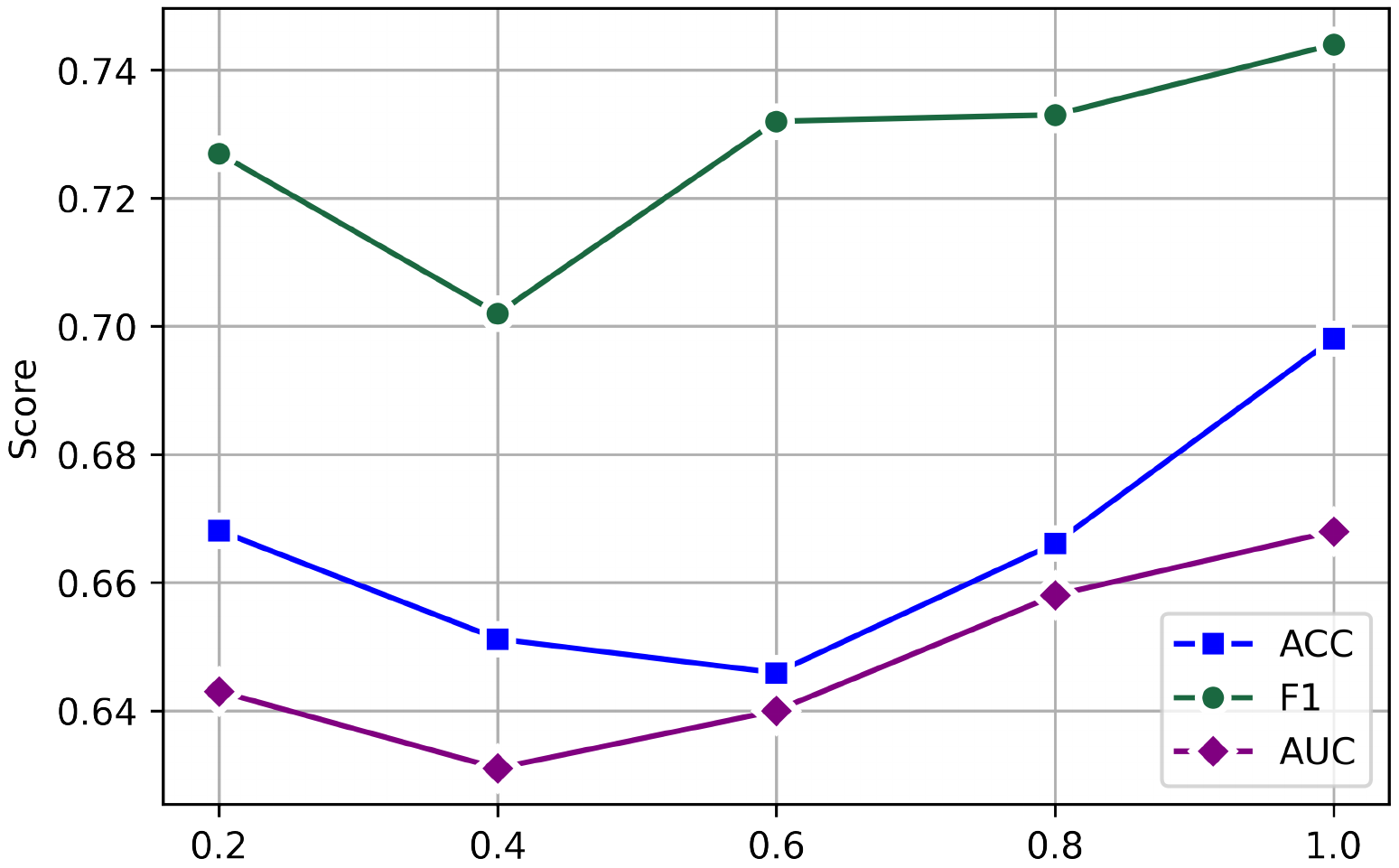}
\vspace{-3pt}
\small (d) ADHD200 Dataset
\end{minipage}

\caption{Evaluation of the loss weight $\alpha$ for $\mathcal{L}_{\text{HiCo}}$. (a) AVG; (b) Tiantan-Tumor Dataset; (c) OASIS-I Dataset; (d) ADHD200 Dataset.}
\label{fig:loss_weight}
\end{figure*}

\subsubsection{Comparison on the ADHD200 datasets}

To rigorously evaluate the cross-site generalizability of our proposed method, we conducted experiments on the multi-center ADHD200~\citep{adhd2012adhd} dataset, which comprises data from six independent imaging sites (KKI, Neuro, NYU, OHSU, Peking, and Pitts). As presented in Table~\ref{tab:ADHD200}, we report the classification accuracy for each individual site, alongside the overall aggregate performance evaluated via three comprehensive metrics. 

Notably, both our single-branch HiCo and dual-branch DBHiCo achieve superior overall performance. Specifically, they yield absolute improvements of at least $0.82\%$ ($68.73\%$ vs. $67.91\%$) and $1.90\%$ ($69.81\%$ vs. $67.91\%$) in average ACC, respectively, compared to the strongest baseline method (VoCo~\citep{wu2024voco}). Furthermore, the site-specific accuracies of our approaches consistently rank as the best or second-best across these highly heterogeneous subsets. It is worth noting that the accuracy on the Pitts subset saturates at $100.0\%$ for the majority of evaluated methods; this anomaly is attributable to the extreme class imbalance within this specific cohort, which consists exclusively of the majority class. In summary, these multi-site evaluation results demonstrate that our method exhibits exceptional robustness and generalizability, effectively mitigating potential domain shifts caused by varying scanning parameters across different institutions.

\begin{figure*}[!t]
\centering
\begin{minipage}[t]{0.24\textwidth}
\centering
\vspace{-3pt}
\includegraphics[width=\linewidth]{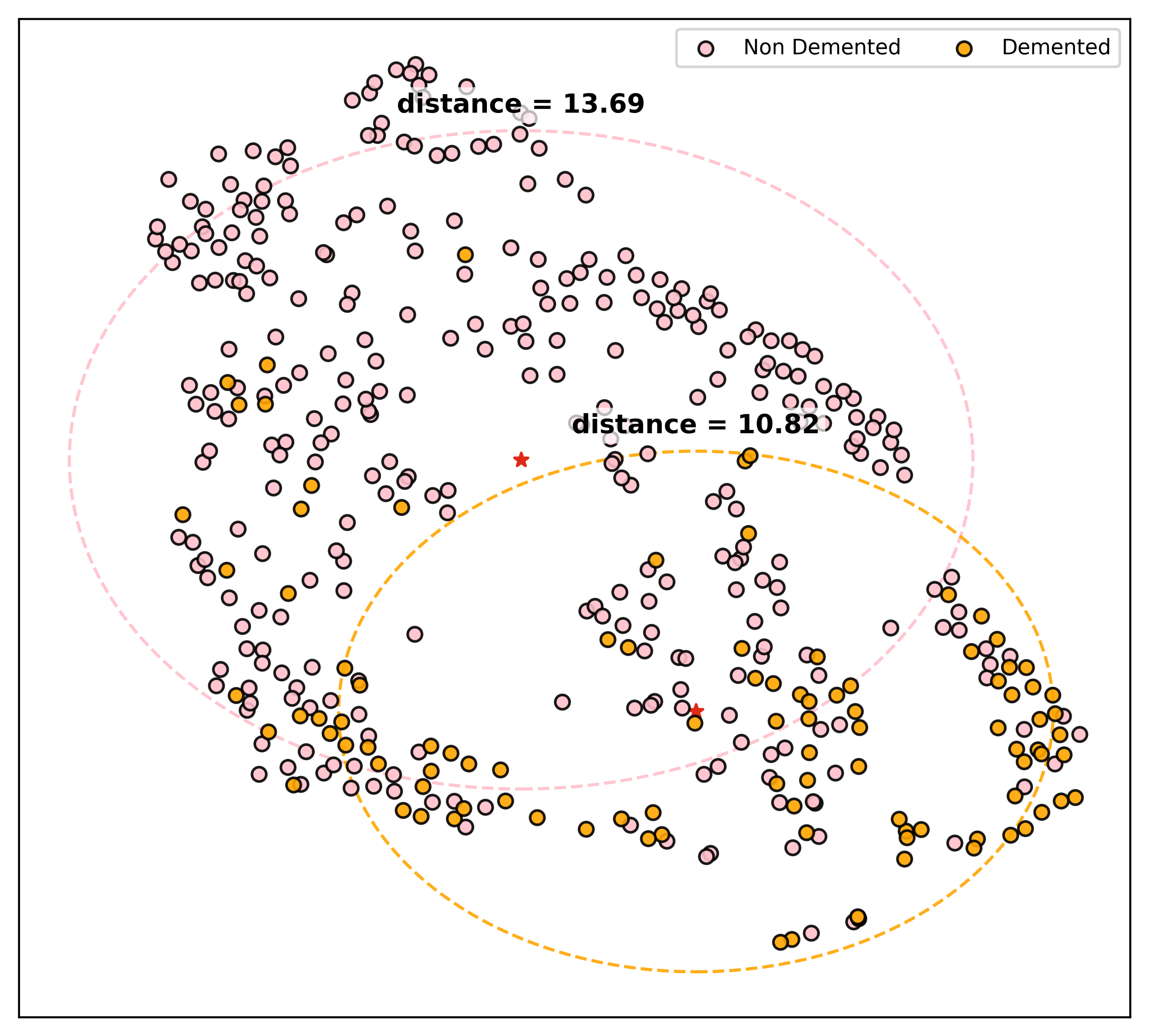}
\vspace{-3pt}
\small (a) SimCLR for OASIS-I
\end{minipage}
\begin{minipage}[t]{0.24\textwidth}
\centering
\vspace{-3pt}
\includegraphics[width=\linewidth]{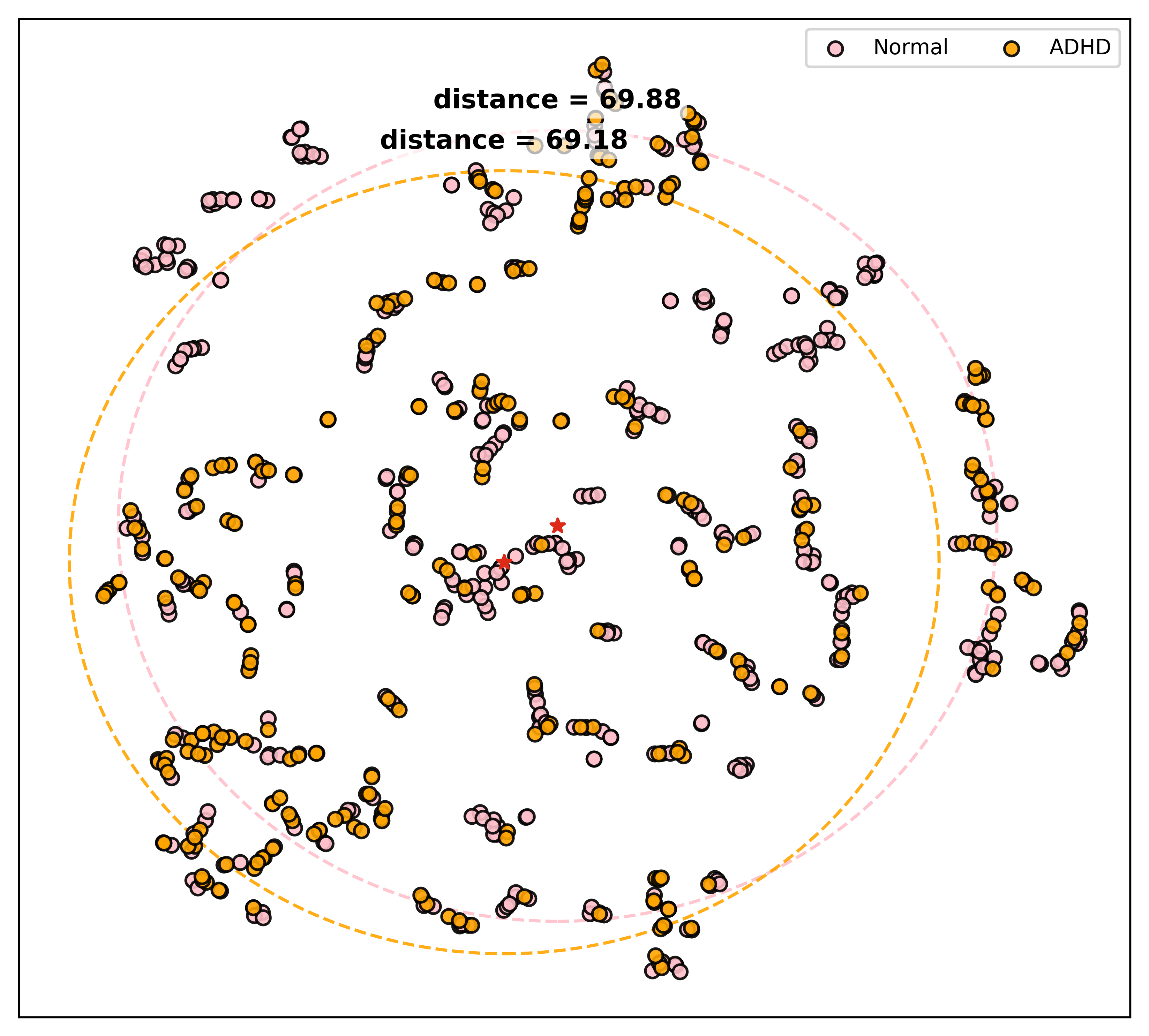}
\vspace{-3pt}
\small (b) SimCLR for ADHD200
\end{minipage}%
\begin{minipage}[t]{0.24\textwidth}
\centering
\vspace{-3pt}
\includegraphics[width=\linewidth]{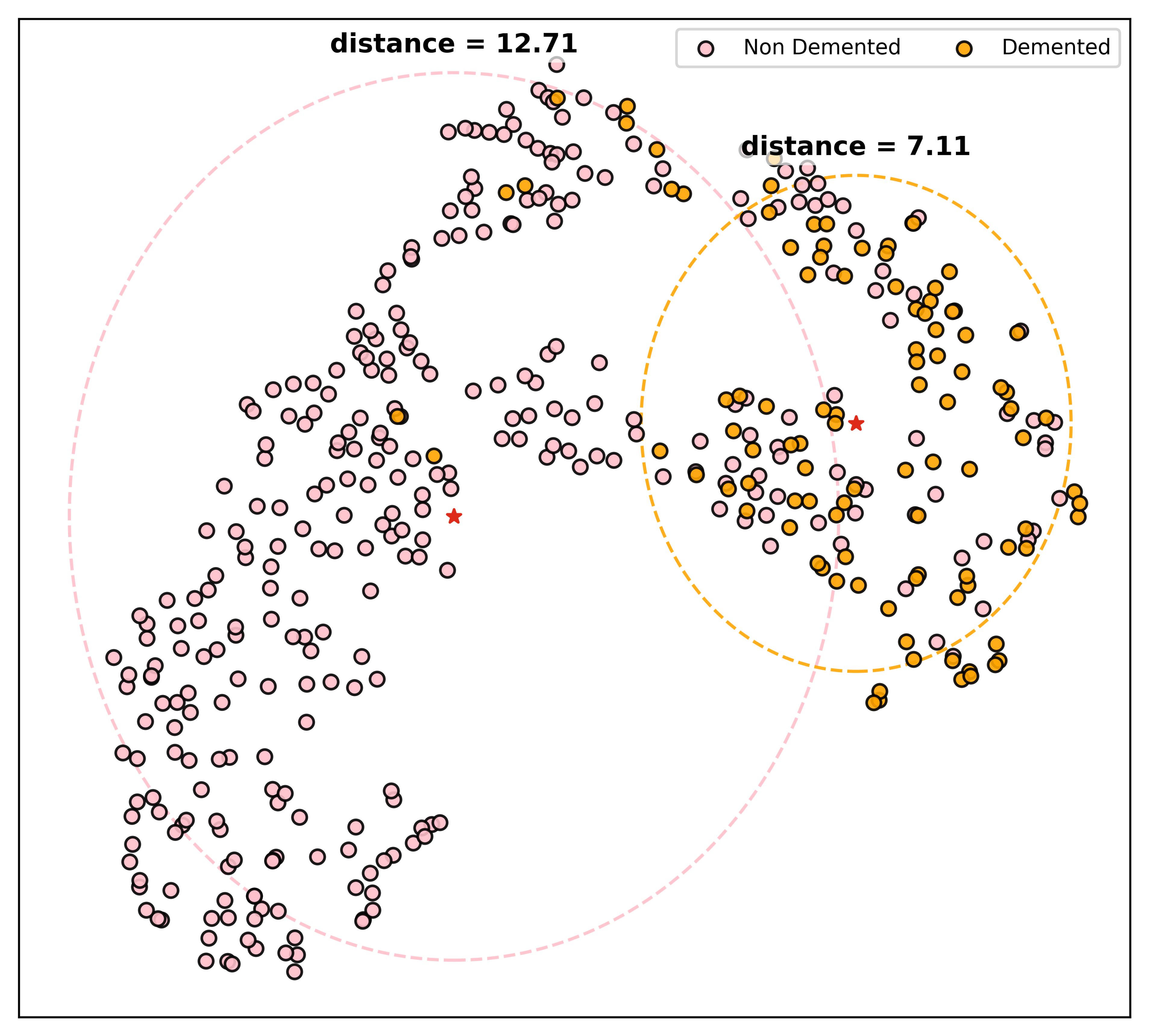}
\vspace{-3pt}
\small (c) DBHiCo for OASIS-I
\end{minipage}%
\begin{minipage}[t]{0.24\textwidth}
\centering
\vspace{-3pt}
\includegraphics[width=\linewidth]{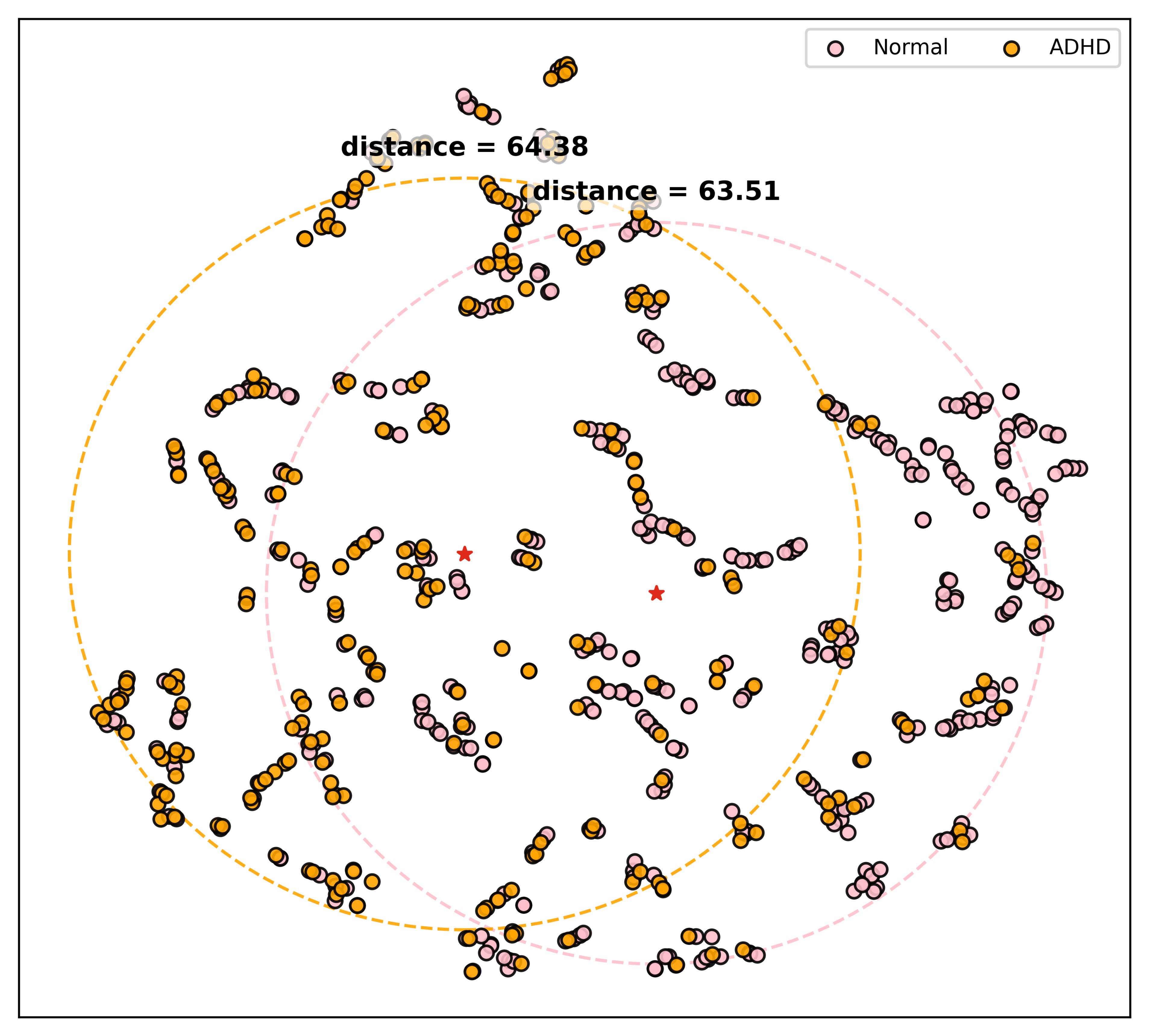}
\vspace{-3pt}
\small (d) DBHiCo for ADHD200
\end{minipage}


\caption{Visualization of T-SNE~\citep{maaten2008visualizing} results. (a)-(b) Feature distributions for SimCLR on OASIS-I and ADHD200, exhibiting noticeable class overlap. (c)-(d) Feature distributions for DBHiCo on the respective datasets, showing enhanced inter-class separability and tighter intra-class compactness.}

\label{fig:vis_tsne}
    \vspace{-5pt}
\end{figure*}

\subsection{Analysis of our proposed method}\label{sec:analysis}
\subsubsection{Ablation study}\label{sec:ablation}
\noindent\textbf{Contrastive views.}\label{sec:ablation_con_type}
As detailed in Section~\ref{sec:construct_pairs}, our DBHiCo leverages original and skull-stripped MRI scans as complementary views for contrastive pair construction. To validate this design, we evaluated several alternative view-construction strategies on the Tiantan dataset (Table~\ref{tab:ablation_views}), including cross-modality contrasting (T1-E mod $\Longleftrightarrow$ T2 mod), varying intensities of data augmentation transformations (Strong trans $\Longleftrightarrow$ Weak trans), and the incorporation of both whole-brain structures and localized tumor tissues (Brain $\Longleftrightarrow$ Tumor). Notably, isolating specific MRI modalities to form contrastive views is the only strategy that degrades performance compared to the non-contrastive baseline, which we attribute to the fact that the joint utilization of all three modalities is indispensable for accurate tumor diagnosis~\citep{rui2025multi}. Ultimately, our proposed `Original $\Longleftrightarrow$ Skull-stripped' paradigm emerges as the optimal strategy, delivering the most substantial performance gains across four out of the five comprehensive evaluation metrics and affirming the value of jointly modeling global anatomical context alongside fine-grained brain tissue features.

\begin{table}[ht]
    \centering
    \caption{Evaluation of model with different contrastive views.}
    \resizebox{1\linewidth}{!}{
        \begin{tabular}{lccccc}
            \whline
            \multirow{2}{*}{Contrastive Types}          & \multicolumn{5}{c}{Tiantan Datasets} \\
            \cmidrule(r){2-6}  
            
                                                        & ACC(\%)             & PRE(\%)             & REC(\%)             & F1(\%)              & AUC(\%)             \\
            \hline
            w/o Dual-Branch \& Contrast                 & $81.88$             & $81.23$             & $77.90$             & $79.18$             & $93.81$             \\
            T1E mod$\Longleftrightarrow$T2 mod          & $81.55$             & $80.09$             & $77.18$             & $77.63$             & $94.56$             \\
            Strong trans$\Longleftrightarrow$Weak trans & $82.38$             & $78.63$             & $79.11$             & $79.40$             & $94.60$             \\
            Brain$\Longleftrightarrow$Tumor             & \underline{$84.27$} & \underline{$83.05$} & \underline{$81.27$} & \underline{$81.92$} & $\mathbf{96.10}$    \\
            Original$\Longleftrightarrow$Skull-stripped & $\mathbf{86.51}$    & $\mathbf{84.69}$    & $\mathbf{84.18}$    & $\mathbf{84.21}$    & \underline{$96.09$} \\
            \whline
        \end{tabular}
    }
    \label{tab:ablation_views}
    \textsuperscript{*}{\ud{\raggedleft{\scriptsize{Note: PRE: Precision, REC: Recall.}}}}
\end{table}

\noindent\textbf{Method components.} \label{sec:ablation_component}
Table~\ref{tab:abaltion_component} assesses the individual and synergistic contributions of DBHiCo's core components: the dual-branch encoder architecture (Dual-Branch), the hierarchical contrastive loss (Hierarchical), and the contrastive projector (Projector). 
The results demonstrate progressive performance gains through cumulative integration, culminating in peak performance across all three datasets when the complete configuration is utilized. Notably, the Hierarchical loss exhibits the most substantial independent impact, yielding a $2.48\%$ absolute accuracy improvement on the ADHD200~\citep{adhd2012adhd} dataset ($68.45\%$ vs. $65.97\%$) even in a single-branch setup. The inclusion of the Projector further refines these representations by mapping features into a more discriminative latent space, providing consistent incremental gains. Furthermore, while the standalone Dual-Branch architecture offers limited benefits over the baseline, it demonstrates a powerful synergistic effect when coupled with the Hierarchical loss and Projector, ultimately driving profound F1-score enhancements of $5.71\%$ ($84.21\%$ vs. $78.50\%$) on Tiantan and $4.51\%$ ($79.86\%$ vs. $75.35\%$) on OASIS-I~\citep{marcus2007open} compared to the isolated dual-branch configuration.

\begin{table}[!ht]
    \centering
    \caption{Evaluation of our DBHiCo's components.}
    \resizebox{1\linewidth}{!}{
        \begin{tabular}{ccccccccc}
            \whline
            \multicolumn{3}{c}{\textbf{Components}}         & \multicolumn{2}{c}{Tiantan} & \multicolumn{2}{c}{OASIS-I} & \multicolumn{2}{c}{ADHD200} \\
            \cmidrule(r){1-3}   \cmidrule(lr){4-5}   \cmidrule(lr){6-7}   \cmidrule(lr){8-9} 
            
            Dual-Branch    & Hierarchical   & Projector      & ACC(\%)             & F1(\%)              & ACC(\%)             & F1(\%)              & ACC(\%)             & F1(\%)              \\ [1pt]
            \hline
            \XSolidBrush   & \XSolidBrush   & \XSolidBrush   & $82.27$             & $79.18$             & $80.56$             & $73.73$             & $65.97$             & $62.03$             \\
            \XSolidBrush   & \CheckmarkBold & \XSolidBrush   & $83.27$             & $80.91$             & $81.98$             & $77.07$             & $68.45$             & $66.66$             \\
            \XSolidBrush   & \CheckmarkBold & \CheckmarkBold & $84.89$             & $82.17$             & $83.17$             & $78.12$             & \underline{$68.73$} & $\mathbf{66.93}$    \\
            \cdashline{1-9}
            \CheckmarkBold & \XSolidBrush   & \XSolidBrush   & $82.16$             & $78.50$             & $81.67$             & $75.35$             & $65.99$             & $63.91$             \\
            \CheckmarkBold & \CheckmarkBold & \XSolidBrush   & \underline{$85.15$} & \underline{$83.01$} & \underline{$83.49$} & \underline{$79.42$} & $68.02$             & $66.32$             \\
            \CheckmarkBold & \CheckmarkBold & \CheckmarkBold & $\mathbf{86.51}$    & $\mathbf{84.21}$    & $\mathbf{85.32}$    & $\mathbf{79.86}$    & $\mathbf{69.81}$    & \underline{$66.78$} \\
            \whline
        \end{tabular}
    }
    \label{tab:abaltion_component}
\end{table}

\noindent\textbf{Loss function.}\label{sec:ablation_backbone}
Table~\ref{tab:ablation_con_loss} demonstrates the empirical necessity of explicitly modeling distinct negative difficulty levels through our proposed hierarchical contrastive loss. When upgrading the standard contrastive loss (detailed in Section~\ref{sec:Preliminary}) to our hierarchical variant, both the single-branch HiCo and dual-branch DBHiCo architectures exhibit consistent and substantial performance gains across all three evaluated datasets. Most notably, integrating this hierarchical objective into the DBHiCo framework yields absolute accuracy improvements of $2.91\%$, $1.99\%$, and $1.69\%$ on the Tiantan, OASIS-I, and ADHD200 benchmarks, respectively, firmly validating that a calibrated negative sampling strategy is crucial for learning highly discriminative medical image representations.

        

\begin{table}[ht]
    \centering
    \caption{Evaluation of $\mathcal{L}_{\text{HiCo}}$ whether using hierarchical difficulty.}
    \resizebox{1\linewidth}{!}{
        \begin{tabular}{lccccccc}
            \whline
            \multirow{2}{*}{Method} & \multirow{2}{*}{Loss function $L_C$}            & \multicolumn{2}{c}{Tiantan} & \multicolumn{2}{c}{OASIS-I} & \multicolumn{2}{c}{ADHD200} \\ [1pt]
            \cmidrule(r){3-4}   \cmidrule(lr){5-6}   \cmidrule(lr){7-8}  
            
                                    &                                                 & ACC(\%)             & F1(\%)             & ACC(\%)             & F1(\%)             & ACC(\%)             & F1(\%)             \\ [-1pt] 
            \hline
            \multirow{2}{*}{HiCo}   & Contrastive loss~\citep{oord2018representation} & $83.72$             & $81.13$            & $82.85$             & $74.88$            & $66.57$             & $64.01$            \\
                                    & Hierarchical Contrastive loss                   & \underline{$84.89$} & \underline{$82.17$} & $83.17$             & \underline{$78.12$} & \underline{$68.73$} & $\mathbf{66.93}$   \\
            \cdashline{1-8}
            \multirow{2}{*}{DBHiCo} & Contrastive loss~\citep{oord2018representation} & $83.60$             & $80.64$            & \underline{$83.33$} & $76.89$            & $68.12$             & $63.72$            \\
                                    &   Hierarchical Contrastive loss &   $\mathbf{86.51}$ &   $\mathbf{84.21}$ &   $\mathbf{85.32}$ &   $\mathbf{79.86}$ &   $\mathbf{69.81}$ &   \underline{$66.78$} \\
            \whline 
        \end{tabular}
    }
    \label{tab:ablation_con_loss}
\end{table}
 
\noindent\textbf{Loss terms of} $\mathcal{L}_{\text{HiCo}}$.\label{sec:ablation_hico_loss}
Next, we systematically evaluate the individual contributions of the four difficulty-aware contrastive constraints ($\mathcal{L}_\mathrm{1}$ to $\mathcal{L}_\mathrm{4}$) formulated from our prior conditions (Section~\ref{sec:construct_pairs}). By isolating each hierarchical level, Table~\ref{tab:ablation_hico_loss} demonstrates that imposing constraints on the most challenging negative pairs generally yields the most substantial performance benefits; for instance, the most stringent condition ($\mathcal{L}_\mathrm{1}$) consistently outperforms the most relaxed condition ($\mathcal{L}_\mathrm{4}$) across all datasets (\textit{e.g.}, $84.09$\% vs. $82.40$\% accuracy on Tiantan). Although the individual effectiveness of these terms progressively diminishes as the negative pair difficulty decreases, integrating all four levels is crucial, as their synergistic combination achieves the absolute peak performance across every metric and benchmark.

\begin{table}[!ht]
    \centering
    \caption{Evaluation of the loss terms $\mathcal{L}_{\text{HiCo}}$.}
    \resizebox{1\linewidth}{!}{%
        \begin{tabular}{cccccccccc}
            \whline
            \multicolumn{4}{c}{\textbf{Loss}} & \multicolumn{2}{c}{Tiantan} & \multicolumn{2}{c}{OASIS-I} & \multicolumn{2}{c}{ADHD200} \\
            \cmidrule(r){1-4}   \cmidrule(lr){5-6}   \cmidrule(lr){7-8}   \cmidrule(lr){9-10}  
            
            $\ud{\mathcal{L}_\mathrm{1}}$ & $\ud{\mathcal{L}_\mathrm{2}}$ & $\ud{\mathcal{L}_\mathrm{3}}$ & $\ud{\mathcal{L}_\mathrm{4}}$ & ACC(\%) & F1(\%) & ACC(\%) & F1(\%) & ACC(\%) & F1(\%) \\
            \hline
            \CheckmarkBold & \XSolidBrush   & \XSolidBrush   & \XSolidBrush   & \underline{$84.09$} & \underline{$81.26$} & \underline{$83.57$} & \underline{$77.65$} & \underline{$66.76$} & \underline{$64.87$} \\
            \XSolidBrush   & \CheckmarkBold & \XSolidBrush   & \XSolidBrush   & $83.72$             & $81.08$             & $82.53$             & $77.29$             & $65.75$             & $63.96$             \\
            \XSolidBrush   & \XSolidBrush   & \CheckmarkBold & \XSolidBrush   & $84.68$             & $82.13$             & $81.82$             & $76.32$             & $65.85$             & $63.58$             \\
            \XSolidBrush   & \XSolidBrush   & \XSolidBrush   & \CheckmarkBold & $82.40$             & $79.53$             & $81.98$             & $77.74$             & $66.23$             & $62.98$             \\
            \CheckmarkBold & \CheckmarkBold & \CheckmarkBold & \CheckmarkBold & $\mathbf{86.51}$    & $\mathbf{84.21}$    & $\mathbf{85.32}$    & $\mathbf{79.86}$    & $\mathbf{69.81}$    & $\mathbf{66.78}$    \\
            \whline
        \end{tabular}
    }
    \label{tab:ablation_hico_loss}
\end{table}

\subsubsection{Hyper-parameter analysis}\label{sec:hyperparameter}

\noindent\textbf{Weights of $\mathcal{L}_{\text{HiCo}}$.}\label{sec:loss_weight}
To evaluate the impact of the hyper-parameter $\alpha$ weighting $\mathcal{L}_{\text{HiCo}}$, we visualize the performance variations across all datasets in Figure~\ref{fig:loss_weight}. The empirical results demonstrate that the evaluation metrics consistently peak when $\alpha$ is set to $1.0$. Generally, increasing the $\alpha$ value up to this optimal threshold yields steady overall performance improvements. This positive trend is particularly pronounced on the more challenging ADHD200~\citep{adhd2012adhd} benchmark, firmly underscoring the critical role of our proposed $\mathcal{L}_{\text{HiCo}}$ in simultaneously reinforcing structural consistency and fine-grained semantic discriminability within MRI representations.

\begin{table}[!ht]
    \centering
    \caption{Evaluation of our DBHiCo's margin-parameter adjustment strategy.}
    \resizebox{1\linewidth}{!}{
        \begin{tabular}{cccccccc}
            \whline
            \multicolumn{2}{c}{\textbf{Margin}} & \multicolumn{2}{c}{Tiantan} & \multicolumn{2}{c}{OASIS-I} & \multicolumn{2}{c}{ADHD200} \\
            \cmidrule(r){1-2}   \cmidrule(lr){3-4}   \cmidrule(lr){5-6}   \cmidrule(lr){7-8}  
            
            Uniform        & Decay          & ACC(\%)                            & F1(\%)                             & ACC(\%)                            & F1(\%)                             & ACC(\%)                            & F1(\%)               \\ [1pt]
            \hline
            \CheckmarkBold & \XSolidBrush   &   $\mathbf{86.51}$ &   $\mathbf{84.21}$ &   $\mathbf{85.32}$ &   $\mathbf{79.86}$ & \underline{$68.26$}                & $65.58$              \\
            \CheckmarkBold & \CheckmarkBold & $85.28$                            & \underline{$83.08$}                & $83.65$                            & $77.84$                            & $66.72$                            & $64.25$              \\
            \XSolidBrush   & \XSolidBrush   & \underline{$85.53$}                & $82.93$                            & $82.94$                            & $76.86$                            &   $\mathbf{69.81}$ &   $\mathbf{66.78}$ \\
            \XSolidBrush   & \CheckmarkBold & $85.05$                            & $82.70$                            & \underline{$84.37$}                & \underline{$79.61$}                & $68.17$                            & \underline{$66.00$}  \\
            \whline
        \end{tabular}
    }
    \label{tab:abaltion_margin}
\end{table}

\noindent\textbf{Margin-parameter adjustment strategy.}\label{sec:ablation_margin} As detailed in Section~\ref{sec:hico_loss}, the margin parameter $m$ is critical for computing $\mathcal{L}_{\text{HiCo}}$. Table~\ref{tab:abaltion_margin} evaluates two adjustment dimensions: \textit{\textbf{Uniform}} (applying a fixed $m=0.1$ across all hierarchical levels versus assigning distinct values of $0.25$, $0.2$, $0.15$, and $0.1$ based on negative difficulty) and \textit{\textbf{Decay}} (maintaining a static margin throughout training versus applying an epoch-wise decay rate of $0.99$). Empirical results indicate that a static, uniform margin strategy optimally balances simplicity and effectiveness on the Tiantan and OASIS-I~\citep{marcus2007open} datasets, achieving peak accuracies of $86.51\%$ and $85.32\%$, respectively. Conversely, for the more challenging and heterogeneous ADHD200~\citep{adhd2012adhd} benchmark, employing distinct, non-decaying margin values yielded the best performance ($69.81\%$ accuracy), demonstrating the flexibility of our methods in tailoring margin constraints to suit varying downstream dataset complexities.

\subsubsection{Interpretability}\label{sec:interpretability}
\noindent\textbf{Grad-CAM visualization.}\label{sec:vis_attn}
We analyze Grad-CAM~\citep{selvaraju2017grad} attention maps across four tumor types of the Tiantan dataset from the perspective of structural-semantic complementarity. As shown in Figure~\ref{fig:vis_attns}, SimCLR mainly focuses on global representation learning, but it is prone to being affected by the background, resulting in blurred attention map boundaries and distorted shapes. For example, in the attention map of Medulloblastoma, the focused area fails to align with the elliptical contour of the cerebellar vermis. In contrast, our HiCo enforces the model to learn fine-grained structural details through hierarchical CL, while DBHiCo further enhances the precise understanding of anatomical structures and pathological tissues.

\noindent\textbf{T-SNE visualization.}\label{sec:vis_tsne}
Furthermore, we analyze T-SNE~\citep{maaten2008visualizing} visualizations by evaluating class separation and centroid distance across OASIS-I~\citep{marcus2007open} and ADHD200~\citep{adhd2012adhd} datasets as shown in Figure~\ref{fig:vis_tsne}. Qualitatively, compared with SimCLR, our DBHiCo shows significantly better discrimination on both of these two binary clusters, indicating that the feature representation of DBHiCo significantly amplifies the inter-class differences. Qualitatively, DBHiCo achieves lower center distances (\textit{e.g.}, on the Demented of OASIS-I~\citep{marcus2007open}, distance for DBHiCo vs. SimCLR: $7.11$ vs. $10.82$), proving that DBHiCo improves the intra-class compactness and thereby achieves more accurate automatic diagnosis.

\section{Conclusion}\label{sec:conclusion}

This paper presents DBHiCo, a novel multi-view representation learning framework designed to overcome the inherent noise and high inter-subject similarity challenges in automated MRI classification. 
Unlike existing contrastive paradigms that rely on rigid binary partitions, DBHiCo collaboratively enforces structural consistency and semantic discrimination through a dedicated dual-branch architecture, seamlessly integrating holistic anatomical boundaries with localized intracranial details derived from paired original and skull-stripped scans. 
To effectively align these fine-grained features, we introduce a hierarchical contrastive loss ($\mathcal{L}_{\text{HiCo}}$) that replaces conventional uniform penalties with a margin-ranked similarity ordering, explicitly calibrating positive and negative pairs based on their cross-view and cross-class difficulty. 
Extensive experiments across three diverse neurological datasets demonstrate that DBHiCo not only establishes state-of-the-art diagnostic performance but also significantly enhances inter-class separability in the latent space.




\section*{Acknowledgements}
This work was supported by the Natural Science Foundation of China under
Grant 62571507.

\bibliographystyle{elsarticle-num}

\bibliography{bibs/references}

\end{document}